\pdfoutput=1

\documentclass[11pt]{article}

\usepackage{naacl2021}

\usepackage{times}
\usepackage{latexsym}

\usepackage{microtype}

\usepackage{hyperref}       
\usepackage{url}            
\usepackage{booktabs}       
\usepackage{amsfonts}       
\usepackage{nicefrac}       
\usepackage{microtype}      

\usepackage{wrapfig,lipsum,booktabs}

\usepackage{amsmath}
\usepackage{algorithm}
\usepackage[noend]{algpseudocode}
\usepackage{multirow}
\usepackage{arydshln}
\usepackage{subcaption}
\usepackage{pgfplots}
\usepackage{tikz}

\usepackage{mathtools}

\newcounter{notecounter}
\newcommand{\enotesoff}{\long\gdef\enote##1##2{}}

\enotesoff

\usepackage{times}
\usepackage{latexsym}

\usepackage[T1]{fontenc}

\usepackage[utf8]{inputenc}

\usepackage{microtype}

\title{
Multi-source Neural Topic Modeling in Multi-view Embedding Spaces
}

\newcommand*{\affaddr}[1]{#1} 
\newcommand*{\affmark}[1][*]{\textsuperscript{#1}}
\newcommand*{\email}[1]{\texttt{#1}}

\renewcommand{\thefootnote}{\fnsymbol{footnote}}
\usepackage{blindtext}
\newcommand\blfootnote[1]{%
\begingroup
\renewcommand\thefootnote{}\footnote{#1}%
\addtocounter{footnote}{-1}%
\endgroup
}

\author{*Pankaj Gupta\affmark[1], *Yatin Chaudhary\affmark[1,2], Hinrich Sch\"{u}tze\affmark[2]\\ 
 \affaddr{\affmark[1]DRIMCo GmbH  Munich, Germany}\\
  \affaddr{\affmark[2]CIS, University of Munich (LMU) Munich, Germany} \\
  \email{info@drimco.net}  | \email{yatin.chaudhary@drimco.net}
}

\begin{document}
\maketitle
\begin{abstract}

Though word embeddings and topics are complementary representations, 
several past works have only used pretrained word embeddings in (neural) topic modeling to address data sparsity
in short-text or small collection of documents. 
This work presents a novel neural topic modeling framework using multi-view embedding spaces:  
(1) pretrained topic-embeddings, and  
(2) pretrained word-embeddings (context-insensitive from Glove and context-sensitive from BERT models)
{\it jointly} from one or {\it many sources} to improve topic quality and better deal with polysemy. 
In doing so, we first build respective pools of pretrained topic (i.e., \texttt{TopicPool}) and word embeddings (i.e., \texttt{WordPool}). We then identify one or more relevant source domain(s) and transfer knowledge to 
guide meaningful learning in the sparse target domain.   
Within neural topic modeling, we quantify the quality of topics and document representations via generalization (perplexity), 
interpretability (topic coherence) and information retrieval (IR) using short-text, long-text, small and large document collections from news and medical domains. 
Introducing the multi-source multi-view embedding spaces, we have shown state-of-the-art neural topic modeling using 6 source (high-resource) and 5 target (low-resource) corpora.

\end{abstract}

\section{Introduction}\label{sec:introduction}
Probabilistic topic models, such as LDA \cite{DBLP:journals/jmlr/BleiNJ03}, Replicated Softmax (RSM) \cite{DBLP:conf/nips/SalakhutdinovH09} and Document Neural Autoregressive Distribution Estimator (DocNADE) \cite{DBLP:conf/nips/LarochelleL12} 
are often used to extract topics from text collections and learn latent document representations 
to perform natural language processing tasks, such as information retrieval (IR).
\blfootnote{* : equal contribution}
Though they have been shown to be powerful in modeling large text corpora,  
the topic modeling (TM) still remains challenging especially in the sparse-data setting,
especially for the cases where word co-occurrence data is insufficient, 
e.g., 
on short-text or a corpus of few documents. 
It leads to a poor quality of topics and representations.

To address data sparsity issues, several works \cite{P15-1077, DBLP:journals/tacl/NguyenBDJ15, pankajgupta:2019iDocNADEe, Guptalifelongicml2020} have introduced external knowledge in traditional topic models, e.g., incorporating 
word embeddings obtained from Glove \cite{D14-1162} or word2vec \cite{Mikolov:word2vec2013}. 
However, no prior work in topic modeling has employed multi-view embedding spaces: (1) {\it pretrained topics}, i.e., topical embeddings obtained from large document collections, and 
(2) {\it pretrained contextualized word embeddings} from large-scale language models like BERT \cite{BERT:DevlinCLT19}. 

\begin{table}[t]
\center
\renewcommand*{\arraystretch}{1.1}
\center
\resizebox{.49\textwidth}{!}{
\begin{tabular}{c|l|c}
\toprule
\multicolumn{1}{c|}{\bf Topic} & \multicolumn{1}{c|}{\bf Topic Words} & \multicolumn{1}{c}{\bf Topic Label} \\ 
\midrule
\multirow{2}{*}{$Z_1$ ($\mathcal{S}^1$)} 	& {profit, growth,  stocks, {\bf apple}, \underline{fall},} &\multirow{2}{*}{\it Trading}\\
& {consumer, buy, billion, shares} & \\ \hline
\multirow{2}{*}{$Z_2$($\mathcal{S}^2$)} 	& {smartphone, ipad, {\bf apple}, app,} & \multirow{2}{*}{\it Product Line}\\
&{iphone, devices, phone, tablet} &  \\ \hline
\multirow{2}{*}{$Z_3$ ($\mathcal{S}^3$)}	& {microsoft, mac, linux, ibm, ios,} & \multirow{2}{*}{\it Operating System}\\
& {{\bf apple}, xp, windows, software} & \\ \hline
\multirow{2}{*}{$Z_4$ ($\mathcal{T}$)}		& {{\bf apple}, talk, computers, {\color{red} shares},} & \multirow{2}{*}{$?$}\\
& {{\color{blue} disease},  driver, electronics, {\color{red} profit}, ios} & \\ \bottomrule	 
\end{tabular}}
\caption{Coherent ({\small $Z_1$-$Z_3$}) vs Incoherent ({\small $Z_4$}) topics 
from high-resource ({\small $S^1$-$S^3$}) and low-resource ({\small $\mathcal{T}$}) texts}
\label{tab:motivation}
\end{table}

Though topics and word embeddings are 
complementary in how they represent the meaning, 
they are distinctive in how they learn from word occurrences observed in text corpora.  
A topic model \citep{DBLP:journals/jmlr/BleiNJ03} 
is a statistical tool to infers topic distributions across a collection of documents 
and assigns a topic to each word occurrence,  
where the assignment is equally dependent on all other words appearing in the same document. 
Therefore, a topic has a {\it global view} representing semantic structures hidden in document collection. 
On other hand,  word embeddings have primarily {\it local view} in the sense that they are learned based on local collocation pattern in a text corpus, 
where the representation of each word often depends on a local context window  
\citep{DBLP:conf/nips/MikolovSCCD13} or is a function of its sentence(s) \citep{N18-1202}. 
Consequently, they are not aware of the thematic structures underlying the document collection. 
Additionally, recent studies \cite{N18-1202, BERT:DevlinCLT19, RoBERTa2019} have shown a reasonable success in several NLP applications by employing
pretrained contextualized word embeddings, 
where the representation of a word is different in different contexts (i.e., context-sensitive).   
In context of this work, the representations due to global and local (context-sensitive or context-insensitive) views together are referred as {\it multi-view} embeddings. 

For example in Table \ref{tab:motivation}, consider four topics ($Z_1$-$Z_4$) of different domains where the topics ($Z_1$-$Z_3$) 
are respectively obtained from three different high-resource source ($\mathcal{S}^1$-$\mathcal{S}^3$) domains whereas $Z_4$ from a 
low-resource target domain $\mathcal{T}$ (especially in the data-sparsity settings). 
Observe that the topics about {\it Trading} ($Z_1$), {\it Product Line} ($Z_2$) and {\it Operating System} ($Z_3$) are coherent and 
and represent meaningful semantics at document-level via lists of topic words. However in sparse-data settings, the topic $Z_4$ 
discovered is incoherent (noisy) and it is difficult to infer meaningful document semantics. 

Unlike the topics, word embeddings (context-insensitive) encode syntactic and semantic relatedness in fine-granularity and therefore, do not capture thematic structures.  
For instance, the top-5 nearest neighbors (NN) of {\it apple} (below) in word embedding \citep{DBLP:conf/nips/MikolovSCCD13} space suggest that it refers to a {\it fruit} and   
do not express any topical information (e.g., {\it Trading}, {\it Product Line} or {\it Health}) in the corpora.
Similarly given the NN of the word {\it fall}, it is difficult to infer its association with document-level semantics, 
e.g., {\it Trading} as expressed by  $Z_1$ in topic-embedding space.  

{\small {{\bf apple} $\xRightarrow{\text{NN}}$ {\it apples, pear, fruit, berry, pears, strawberry}}

{{\bf fall} $\xRightarrow{\text{NN}}$ {\it falling, falls, drop, tumble, rise, plummet}}}
{

Therefore, topic and word embedding spaces encode complementary semantics. 
Different to context-insensitive word embeddings, the word {\it apple} is referring to an organization and contextualized by different topical semantics respectively 
in the three sources {$\mathcal{S}^1$-$\mathcal{S}^3$}. Thus, it arises the need for context-sensitive embeddings in topic modeling.  

\begin{table}[t]
\center
\resizebox{.45\textwidth}{!}{
\begin{tabular}{c|l}
\multicolumn{1}{c|}{\bf Notation} & \multicolumn{1}{c}{\bf Description} \\ 
\midrule
LVT, GVT  &  Local-view Transfer, Global-view Transfer  						\\ 
MVT, MST &   Multi-view Transfer, Multi-source Transfer 		 			\\
$\mathcal{T}$, $\mathcal{S}$  & A target domain, a set of source domains  		\\
${\bf v}$, $k$, $\mathcal{L}$  & An input document, $k$th source, loss \\
$ K, D$  & 	Vocabulary size, document size			 \\
$E$,  $H$ & Word embedding dimension, \#topics\\
${\bf W} \in \mathbb{R}^{H \times K}$   & Encoding matrix of DocNADE  in $\mathcal{T}$ \\
${\bf U} \in \mathbb{R}^{K \times H}$   & Decoding matrix of DocNADE \\
$ \lambda^{k}$ 	 & Degree of relevance of ${\bf E}^{k}$ in $\mathcal{T}$ 	\\
$ \gamma^{k}$  &	Degree of imitation of ${\bf Z}^{k}$ by ${\bf W}$			  \\
${\bf E}^{k} \in \mathbb{R}^{E \times K}$   & Word embeddings of $k$th source \\ 
${\bf Z}^{k} \in \mathbb{R}^{H \times K}$   & Topic embeddings of $k$th source  \\ 
${\bf A}^{k} \in \mathbb{R}^{H \times H}$    & Topic-alignment in $\mathcal{T}$ and ${\bf Z}^{k}$ \\
${\bf b} \in \mathbb{R}^{K}$, ${\bf c} \in \mathbb{R}^{H}$   & Visible-bias, hidden-bias \\
$DC$ & Document Collection
\end{tabular}}
\caption{Description of the notations used in this work}
\label{tab:notations}
\end{table} 
\begin{figure*}[t]
  \centering
\includegraphics[scale=.64]{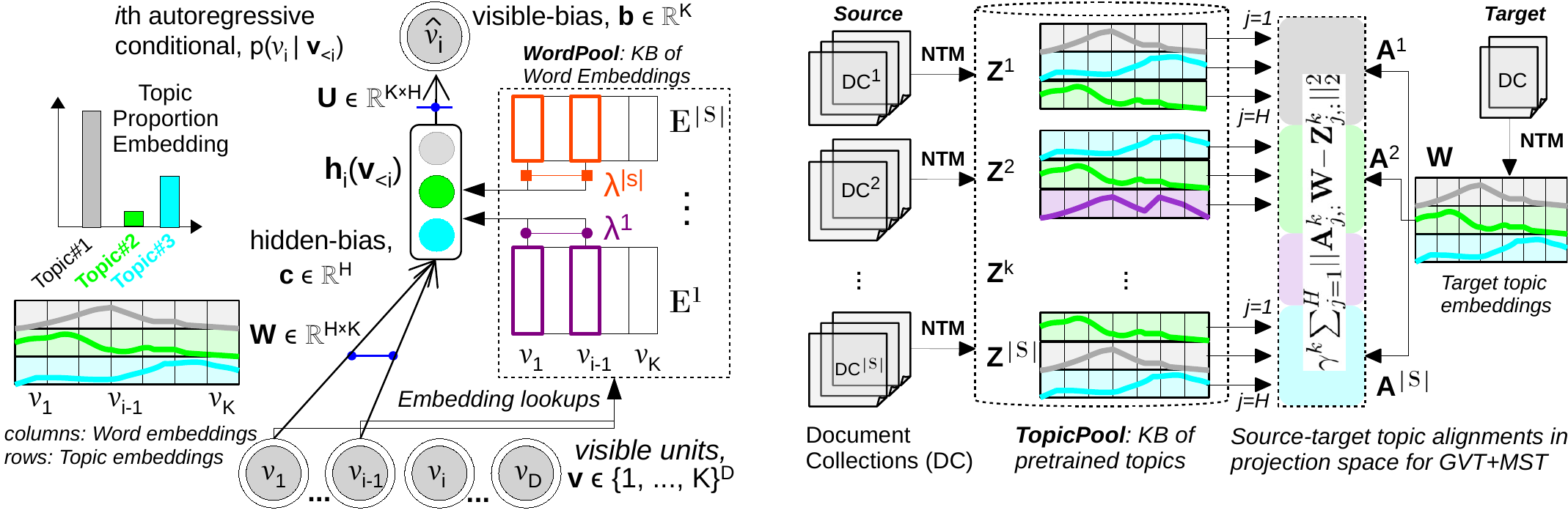}
  \caption{(Left) DocNADE (LVT+MST): Multi-source transfer learning in NTM for a document ${\bf v}$ by introducing pretrained word embeddings
 from a \texttt{WordPool} at each autoregressive step $i$.
Double circle $\rightarrow$ multinomial (softmax) unit  \citep{DBLP:conf/nips/LarochelleL12}. 
(Right) DocNADE (GVT+MST): Multi-source transfer learning in NTM by introducing pretrained (latent) topic embeddings from a \texttt{TopicPool}, 
illustrating topic alignments between source and target corpora. 
Each outgoing row from ${\bf Z}^{k}$$\in$$\mathbb{R}^{H \times K}$ signify a topic embedding 
of corresponding $k$th source corpus, $DC^{k}$.  
Here, NTM refers to a DocNADE \citep{DBLP:conf/nips/LarochelleL12} based Neural Topic Model. 
}
  \label{fig:AutoregressiveNetworks}
\end{figure*}

{\bf Contribution (1) Multi-view Neural Topic Modeling using pretrained word and topic embeddings}:  
To alleviate the data sparsity issues, 
it is the {\bf first work} in unsupervised neural topic modeling (NTM) within transfer learning paradigm that employs {\it multi-view embedding spaces} via: 
(a) {\it Global-view Transfer} ({\it GVT}): Pretrained topic embeddings instead of using word embeddings exclusively, and 
(b) {\it Multi-view Transfer} ({\it MVT}): Pretrained  topic and word embeddings (context-insensitive from Glove \cite{D14-1162} and 
context-sensitive from large-scale language models such as BERT \citep{BERT:DevlinCLT19} 
{\it jointly} to address data sparsity and polysemy issues. 

{\bf Contribution (2) Multi-source Multi-view Neural Topic Modeling}:  
A single source of prior knowledge is often insufficient due to incomplete and non-overlapping domain information required by a target domain.   
Therefore, there is a need to leverage multiple sources of prior knowledge, dealing with domain-shifts \citep{DBLP:conf/aaai/CaoPZYY10} among the target and sources. 
In doing so, we first learn word and topic representations on multiple source domains to build \texttt{WordPool} and \texttt{TopicPool}, respectively and then perform {\it multi-view} and {\it multi-source} 
transfer learning in neural topic modeling by jointly using the complementary representations. 

We evaluate the effectiveness of multi-source neural topic modeling in multi-view embedding spaces  
using 7 (5 low-resource and 2 high-resource) target and 5 (high-resource) source corpora from news and medical domains, consisting of  short-text, long-text, small and large document collections. 
We have shown state-of-the-art results with significant gains quantified 
by generalization (perplexity), interpretability (topic coherence) and text retrieval. 
The code is available at \url{https://github.com/YatinChaudhary/Multi-view-Multi-source-Topic-Modeling}.

\section{Knowledge-Aware Topic Modeling}
Consider a sparse target domain $\mathcal{T}$ and a set of source domains  $\mathcal{S}$, 
we first prepare two knowledge bases (KBs) of representations (or embeddings) from document collections of each of the  $|\mathcal{S}|$ sources: 
(1) {\texttt{WordPool}}: a KB of pretrained word embeddings matrices $\{ {\bf E}^1, ..., {\bf E}^{|\mathcal{S}|} \}$, where ${\bf E}^k \in \mathbb{R}^{E \times K}$,  and 
(2) {\texttt{TopicPool}}: a KB of pretrained latent topic embeddings $\{ {\bf Z}^1, ..., {\bf Z}^{|\mathcal{S}|} \}$, 
where ${\bf Z}^k \in \mathbb{R}^{H \times K}$ encodes a distribution over a vocabulary of $K$ words. Here, 
$k \in [1,...,|\mathcal{S}|]$ in superscript indicates knowledge of $k$th source, and  $E$ and $H$ are word embedding and latent topic dimensions, respectively.  
While topic modeling on $\mathcal{T}$, we introduce the two types of knowledge transfers from one or many sources: 
{\it  Local} (\texttt{LVT}) and {\it Global} (\texttt{GVT}) {\it View Transfer} using the two KBs of pretrained word (i.e., \texttt{WordPool}) and topic (i.e., \texttt{TopicPool}) embeddings, respectively.   
Specially, we employ a neural autoregressive topic model, i.e., DocNADE as backbone in building  the pools 
and realizing the multi-source multi-view framework. 

Table \ref{tab:notations} describes the notations used. 
{\it Notice} that the superscript used in notations indicates a source. 

\subsection{Neural Autoregressive Topic Models} 
DocNADE \citep{DBLP:conf/nips/LarochelleL12} is an unsupervised neural-network based generative topic model that is inspired by the 
benefits of NADE \citep{DBLP:journals/jmlr/LarochelleM11} and RSM \citep{DBLP:conf/nips/SalakhutdinovH09} architectures. 
Specifically, DocNADE factorizes the joint probability distribution of words in a document
as a product of conditional distributions and efficiently models each conditional via
a feed-forward neural network (ff-net), following reconstruction mechanism.  



{\bf DocNADE Formulation}: 
For a document ${\bf v}$ = $(v_1, ..., v_D)$ of size $D$, each word index $v_i$ takes value in $\{1, ..., K\}$ of vocabulary size $K$.  
DocNADE learns topics 
 in a language modeling fashion \citep{DBLP:journals/jmlr/BengioDVJ03} and 
decomposes the joint distribution $p({\bf v})$=$\prod_{i=1}^{D} p(v_i | {\bf v}_{<i})$ such that 
each autoregressive conditional $p(v_i | {\bf v}_{<i})$ is modeled by a ff-net 
using preceding words ${\bf v}_{<i}$ in the sequence:\vspace{-0.3cm}

{\small \begin{equation*}\label{eq:DocNADEconditionals}
\begin{split}
{\bf h}_i({\bf v}_{<i})   =  g ( {\bf c} + &  \sum_{q<i} {\bf W}_{:, v_q})  \ \ \mbox{and} \ \  g = \{\mbox{sigmoid}, \mbox{tanh}\} \\ 
p (v_i = w | {\bf v}_{<i})  & = \frac{\exp (b_w + {\bf U}_{w,:} {\bf h}_i ({\bf v}_{<i}))}{\sum_{w'} \exp (b_{w'} + {\bf U}_{w',:} {\bf h}_i ({\bf v}_{<i}))}
\end{split}
\end{equation*}}
for each word $i \in \{1,...,D\}$ where ${\bf v}_{<i}$ is the subvector consisting of all $v_q$ such that $q < i$ i.e., ${\bf v}_{<i} \in \{v_1, ...,v_{i-1}\}$, 
$g(\cdot)$ is a non-linear activation function, ${\bf W} \in \mathbb{R}^{H \times K}$ and ${\bf U} \in \mathbb{R}^{K \times H}$ 
are weight 
matrices, ${\bf c} \in \mathbb{R}^H$ and ${\bf b} \in \mathbb{R}^K$ are bias parameter 
vectors. $H$ is the number of hidden units 
(the number of topics to be discovered). 

\begin{algorithm}[t]
\centering
\small 
\caption{{\small 
Computation of $\log p({\bf v})$ and Loss $\mathcal{L}({\bf v})$}}\label{algo:computelogpv} 
{
\begin{algorithmic}[1]
\Statex \textbf{Input}: Source domains ${\mathcal{S}}$, a target domain ${\mathcal{T}}$
\Statex \textbf{Input}: A training document ${\bf v}$ from ${\mathcal{T}}$
\Statex \textbf{Input}: \texttt{WordPool}: A KB of pretrained word embedding matrices $\{{\bf E}^{1}, ..., {\bf E}^{\mathcal{|S|}}\}$ from ${\mathcal{S}}$ domains
\Statex \textbf{Input}: \texttt{TopicPool}: A KB of pretrained latent topics $\{{\bf Z}^{1}, ..., {\bf Z}^{\mathcal{|S|}}\}$ from ${\mathcal{S}}$ domains
\Statex \textbf{Parameters}: ${\bf \Theta} = \{{\bf b}, {\bf c}, {\bf W}, {\bf U},  {\bf A}^{1}, ...,  {\bf A}^{\mathcal{|S|}}, {\bf P}\}$
\Statex \textbf{Hyper-params}: ${\bf \Phi}$ = $\{\lambda^{1},..., \lambda^{\mathcal{|S|}},$ $\gamma^{1},..., \gamma^{\mathcal{|S|}}, H\}$
\State Initialize: ${\bf a} \gets {\bf c}$ and $ p({\bf v}) \gets 1$
\For {word $i$ from $1$ to $D$}
        \State Compute $i^{th}$ position-dependent hidden:
        \Statex  $\qquad$ ${\bf h}_{i} ({\bf v}_{<i})  \gets g({\bf a}$), where $g$ = \{sigmoid, tanh\}
        \State Compute $i^{th}$  autoregressive conditional:
        \Statex $\qquad$ $p(v_{i}=w | {\bf v}_{<i}) \gets \frac{\exp (b_w + {\bf U}_{w,:} {\bf h}_i ({\bf v}_{<i}))}{\sum_{w'} \exp (b_{w'} + {\bf U}_{w',:} {\bf h}_i ({\bf v}_{<i}))}$
	\State  Memorize: $p({\bf v}) \gets  p({\bf v}) p(v_{i} | {\bf v}_{<i})$
       \State Compute pre-activation for word $i$ : 
       \Statex $\qquad$ ${\bf a} \gets {\bf a} + {\bf W}_{:, v_{i}}$  
        \If  {\texttt{LVT}} 
             \State Get word-embeddings ${\bf E}$ from \texttt{WordPool}
             \State Introduce prior knowledge ${\bf E}$ for word $i$:
             \Statex $\qquad \quad \ \ $  {\it scheme} (i): ${\bf a} \gets {\bf a} + \sum_{k=1}^{\mathcal{|S|}} \lambda^k \ {\bf E}_{:, v_{i}}^{k}$
		\Statex $\qquad \quad \ \ $  {\it scheme} (ii): ${\bf \hat e}_i \gets \mbox{concat}({\bf E}_{:, v_{i}}^{1},..., {\bf E}_{:, v_{i}}^{k})$
		\Statex $\qquad \qquad \qquad \qquad \ \ $ ${\bf a} \gets {\bf a} + {\bf P} \cdot {\bf \hat e}_i$
        \EndIf
\EndFor
\State Loss (negative log-likelihood): $\mathcal{L}({\bf v}) \gets  - \log p({\bf v})$
\If  {\texttt{GVT}} 
         \State Topic-embedding transfer using \texttt{TopicPool}: 
         \Statex $\qquad$ $\Delta \gets \sum_{k=1}^{\mathcal{|S|}} \gamma^k  \ \sum_{j=1}^{H}  ||  {\bf A}_{j,:}^{k} {\bf W} - {\bf Z}_{j, :}^{k} ||_2^2$
	  \State Overall loss with controlled topic-imitation: 
         \Statex $\qquad$ $\mathcal{L}({\bf v}) \gets \mathcal{L}({\bf v})  + \Delta$
\EndIf
\State Minimize  $\mathcal{L}({\bf v})$ using stochastic gradient descent 
\end{algorithmic}}
\end{algorithm}

Figure \ref{fig:AutoregressiveNetworks} (left) (except \texttt{WordPool}) describes the DocNADE architecture for 
the $i$th autoregressive step, where 
the parameter $\bf W$ is shared in the feed-forward networks and ${\bf h}_i$ encodes latent document-topic proportion. 
The value of each unit $j$ in the hidden vector signifies contribution of the $j$th topic in the proportion.  
Importantly, the topic-word matrix $\bf W$ has a property that  
the column vector ${\bf W}_{:, v_i}$ corresponds to embedding of the word $v_i$, 
whereas the row vector ${\bf W}_{j, :}$ encodes latent features for 
the $j$th topic (i.e., topic-word distribution).  
We leverage this property to introduce external knowledge via word and topic embeddings. 

Algorithm \ref{algo:computelogpv} (for DocNADE, set both \texttt{LVT} and \texttt{GVT} to {\it False})
demonstrates the computation of $\log p({\bf v})$ and loss (i.e., negative log-likelihood) 
$\mathcal{L}({\bf v})$ 
that is minimized using stochastic gradient descent. 
Moreover, computing each ${\bf h}_i$ is efficient (linear complexity)
due to NADE architecture that leverages the pre-activation ${\bf a}_{i-1}$ of $(i-1)$th
step in computing 
${\bf a}_i$ for the $i$th step (line \#6). 
See \citet{DBLP:conf/nips/LarochelleL12} for further details.  

{\it Why DocNADE backbone}: It 
has shown outperforming traditional models such as LDA 
and RSM. 
Additionally, \citet{pankajgupta:2019iDocNADEe, GuptatexttovecICLR2019} have extended DocNADE on short texts by introducing 
context-insensitive word embeddings; 
however, based on a single-source transfer. 
Thus, we adopt DocNADE. 
  

\subsection{MVT and MST in Neural Topic Modeling}
We describe our transfer learning framework in topic modeling that jointly exploits the complementary prior knowledge 
accumulated in (\texttt{WordPool}, \texttt{TopicPool}),  
obtained from large document collections (DCs) from several sources.  
In doing so, we first apply the DocNADE to generate a topic-word matrix for each of the DCs, 
where its column-vector and row-vector generate ${\bf E}^{k}$ and ${\bf Z}^{k}$, respectively for the $k$th source. 
See 
{\it appendix} 
for the mechanics of extracting word and topic embeddings from the topic-word matrix of a source. 

\begin{table*}[t]
    \begin{minipage}{.76\linewidth}
      \center
	\renewcommand*{\arraystretch}{1.1}
	\resizebox{.995\linewidth}{!}{
       \begin{tabular}{r|r|rrrrrr||r|r|rrrrrr}
\toprule
\multicolumn{8}{c||}{\texttt{Target Domain Corpora}} & \multicolumn{8}{c}{\texttt{Source Domain Corpora}} \\ 
 \multicolumn{1}{c|}{\bf ID} & \multicolumn{1}{c|}{\bf Data} &  \multicolumn{1}{c}{\bf Train} &  \multicolumn{1}{c}{\bf Val} & \multicolumn{1}{c}{\bf Test} &  \multicolumn{1}{c}{$K$}   & \multicolumn{1}{c}{\bf L} & \multicolumn{1}{c||}{\bf C}     
&   \multicolumn{1}{c|}{\bf ID} &  \multicolumn{1}{c|}{\bf Data} &  \multicolumn{1}{c}{\bf Train} &  \multicolumn{1}{c}{\bf Val} & \multicolumn{1}{c}{\bf Test} &  \multicolumn{1}{c}{$K$}   & \multicolumn{1}{c}{\bf L} & \multicolumn{1}{c}{\bf C}    \\ \midrule 
$\mathcal{T}^1$	&    20NSshort             & 1.3k       & 0.1k        & 0.5k           &    1.4k           &  13.5        &   20                   	&  $\mathcal{S}^1$  	& 20NS                  & 7.9k & 1.6k & 5.2k  &     2k          &   107.5     &   20        \\
$\mathcal{T}^2$	&    20NSsmall             & 0.4k      &  0.2k       & 0.2k            &      2k             &   187.5     &    20                 	&  $\mathcal{S}^2$  	&  R21578              &  7.3k&  0.5k & 3.0k  &      2k               &   128       &   90       \\
$\mathcal{T}^3$	&    TMNtitle                & 22.8k    &  2.0k     & 7.8k              &       2k           &   4.9          &     7                  	&  $\mathcal{S}^3$  	&   TMN                 & 22.8k  &  2.0k & 7.8k  &      2k         &    19          &       7     \\
$\mathcal{T}^4$	&    R21578title            &  7.3k     &  0.5k     & 3.0k            &   2k                  &   7.3          &     90               	&  $\mathcal{S}^4$  	& AGNews              & 118k & 2.0k &   7.6k &        5k         &  38       &     4         \\
$\mathcal{T}^5$	&    Ohsumedtitle         &  8.3k     &  2.1k     & 12.7k            &   2k                  &   11.9          &     23               &  $\mathcal{S}^5$  & PubMed             &  15.0k     &  2.5k     & 2.5k            &   3k                  &   254.8          &     -    \\
$\mathcal{T}^6$  	& Ohsumed          &  8.3k     &  2.1k     & 12.7k            &   3k                  &   159.1          &     23 			&              &       &      &           &                     &            &        \\  	   \bottomrule          
\end{tabular}}
\caption{Data statistics: Short/long texts and/or small/large corpora in target and source domains.  
Symbols-  $K$: vocabulary size,  $L$: average text length (\#words), $C$: \#classes and  $k$: thousand. 
For short-text,  $L$$<$$15$. $\mathcal{S}^3$ is also used in target. `-': unlabeled data.}
\label{tab:datadescription}
    \end{minipage} \ \ %
\begin{minipage}{.22\linewidth}
      \centering
	\renewcommand*{\arraystretch}{1.2}
	\resizebox{.995\linewidth}{!}{
        \begin{tabular}{c||c|c|c|c|c|c} 
				& $\mathcal{T}^1$	& $\mathcal{T}^2$ 	& $\mathcal{T}^3$ 	& $\mathcal{T}^4$ 	& $\mathcal{T}^5$ 	 & $\mathcal{T}^6$  \\ \hline \hline
$\mathcal{S}^1$	& $\mathcal{I}$ 			& $\mathcal{I}$  			& $\mathcal{R}$ 			& $\mathcal{D}$  			&  $\mathcal{D}$ 	  		&  $\mathcal{D}$	  \\ \hline
$\mathcal{S}^2$	& $\mathcal{D}$  			& $\mathcal{D}$ 			& $\mathcal{D}$  			& $\mathcal{I}$  			&  $\mathcal{D}$ 	  		&  $\mathcal{D}$	  \\ \hline
$\mathcal{S}^3$	& $\mathcal{R}$  			& $\mathcal{R}$ 			& $\mathcal{I}$				& $\mathcal{D}$ 			&  $\mathcal{D}$ 	  		&  $\mathcal{D}$	  \\ \hline
$\mathcal{S}^4$	& $\mathcal{R}$ 			& $\mathcal{R}$ 			& $\mathcal{R}$ 			& $\mathcal{D}$			&  $\mathcal{D}$			&  $\mathcal{D}$ 	  \\ \hline
$\mathcal{S}^5$	& $\mathcal{D}$  			& $\mathcal{D}$ 			& $\mathcal{D}$ 			& $\mathcal{D}$			& {-}		  		& {-} 	  
\end{tabular}}
\caption{Domain overlap in source-target corpora. $\mathcal{I}$: Identical,  $\mathcal{R}$: Related and $\mathcal{D}$: Distant domains.}
\label{tab:domainoverlap}
    \end{minipage} 
\end{table*}

{\bf LVT+MST Formulation for Multi-source Word Embedding Transfer}:  
As illustrated in Figure \ref{fig:AutoregressiveNetworks} (left) and Algorithm \ref{algo:computelogpv} (with \texttt{LVT} being {\it True}, line \#7), 
we perform transfer learning on a target $\mathcal{T}$ using the \texttt{WordPool} of pretrained word embeddings $\{ {\bf E}^1, ..., {\bf E}^{|\mathcal{S}|} \}$ 
from several sources $\mathcal{S}$ (i.e., multi-source) under the two schemes:

\textbf{\textit{scheme} (i)}: Using a domain-relevance factor $\lambda$ for every source in the \texttt{WordPool} such that the hidden vector ${\bf h}_i$ 
encodes document-topic distribution, augmented with prior knowledge in form of pretrained word embeddings from several sources:\vspace{-0.3cm} 

{\small \begin{equation*}\label{eq:LVTMST}
{\bf h}_i({\bf v}_{<i})  =  g ({\bf c} + \sum_{q<i} {\bf W}_{:, v_q} +  \sum_{q<i} \sum_{k=1}^{\mathcal{|S|}} \lambda^k \ {\bf E}_{:, v_{q}}^{k} ) 
\end{equation*}}
Here, $k$ refers to the $k$th source and 
$\lambda^{k}$ is a weight for 
${\bf E}^{k}$ that 
controls the amount of knowledge transferred in $\mathcal{T}$, based on cross-domain overlap. 

\textbf{\textit{scheme} (ii)}: Using a projection matrix ${\bf P} \in \mathbb{R}^{H \times P}$ with $P=E \times |\mathcal{S}|$ in order to align word-embedding spaces  
of the target and all source domains for all $D$ words in the document ${\bf v}$ such that:\vspace{-0.3cm}
 
{\small \begin{equation*}\label{eq:LVTMSTproject}
\begin{split}
\mbox{For } q \in  \{i,...D\}:  {\bf \hat e}_{q}  =  \mbox{concat}&({\bf E}_{:, v_{q}}^{1},..., {\bf E}_{:, v_{q}}^{k})\\
{\bf h}_i({\bf v}_{<i})  =  g ({\bf c} + \sum_{q<i} {\bf W}_{:, v_q} & + \sum_{q<i}  {\bf P} \cdot {\bf \hat e}_{q})
\end{split}
\end{equation*}} 
Unlike scheme (i), the second schema allows us to automatically determine shifts in the target and source domains, identify and transfer relevant prior knowledge 
from many sources without 
configuring $\lambda$ for every source.   
To better guide TM, we also introduce pre-trained contextualized word embedding from BERT, concatenating with ${\bf \hat e}_{q}$. 

{\bf GVT+MST Formulation for Multi-source Topic Embedding Transfer}:  
Next, we perform knowledge transfer exclusively using the \texttt{TopicPool} of pretrained topic embeddings (e.g., ${\bf Z}^k$) from one or several sources, $\mathcal S$. 
In doing so, we add a regularization term to the loss function $\mathcal{L}({\bf v})$ and require 
DocNADE to minimize the overall loss in a way that the (latent) topic features in  ${\bf W}$
simultaneously inherit relevant topical features from each of the source domains $\mathcal{S}$, 
and thus, it generates meaningful representations for the target $\mathcal{T}$ in order to address data-sparsity. 
The overall loss $\mathcal{L}({\bf v}) $ due to GVT+MST configuration in DocNADE is:\vspace{-0.3cm}

{\small \begin{equation*}
\mathcal{L}({\bf v}) =  - \log p({\bf v}) + \sum_{k=1}^{\mathcal{|S|}} \gamma^k  \ \sum_{j=1}^{H}  ||  {\bf A}_{j,:}^{k} {\bf W} - {\bf Z}_{j, :}^{k} ||_2^2
\end{equation*}}
Here, ${\bf A}^{k}$$\in$$\mathbb{R}^{H \times H}$ aligns latent topics in  the target $\mathcal{T}$ and $k$th source,  
and $\gamma^{k}$ governs the degree of imitation of topic features ${\bf Z}^{k}$ 
by ${\bf W}$ in $\mathcal{T}$. 
Consequently, the generative process of learning meaningful topics in ${\bf W}$ of the target domain ${\mathcal T}$
is guided by relevant topic features $\{{\bf Z}\}_{1}^{\mathcal{|S|}} \in \texttt{TopicPool}$. 
Algorithm \ref{algo:computelogpv} (line \#11) describes the computation of the loss, when \texttt{GVT} = {\it True} and \texttt{LVT} = {\it False}. 

Moreover, Figure \ref{fig:AutoregressiveNetworks} (right) illustrates the need for topic alignments between target and source(s). 
Here, $j$ indicates the topic (i.e., row) index in a topic matrix, e.g., ${\bf Z}^{k}$.   
Observe that the first topic (gray curve), i.e., $Z_{j=1}^{1} \in {\bf Z}^{1}$ of the first source aligns with the first row-vector (i.e., topic) of ${\bf W}$ (of target).  
However, the other two topics  $Z_{j=2}^{1},  Z_{j=3}^{1} \in {\bf Z}^{1}$ need alignment with the target. 
 

{\bf MVT+MST Formulation for Multi-source Word and Topic Embeddings Transfer}:  
When \texttt{LVT} and \texttt{GVT} are {\it True} (Algorithm \ref{algo:computelogpv}) for many sources, 
the two complementary representations are jointly used in transfer learning using \texttt{WordPool} and \texttt{TopicPool}, 
 and therefore, the name {\it multi-view} and {\it multi-source} transfers.

{\bf Computational complexity of NTM}: 
For DocNADE, the complexity of computing all hidden layers ${\bf h}_i({\bf v}_{<i})$ is in $O(DH)$ and all $p({\bf v}| {\bf v}_{<i})$ in $O(KDH)$. Thus, the overall complexity of DocNADE is in $O(DH + KDH)$.

Within the proposed transfer learning framework, the complexity of computing all hidden layers (\texttt{LVT+MST} in {\it scheme} (i))  and 
topic-embedding transfer term (\texttt{GVT+MST}) is in  $O(DH+ |\mathcal{S}| DH)$ and  $O(|\mathcal{S}| K H)$, respectively.  
Since $|\mathcal{S}|$$<<$$H$, thus the overall complexity of DocNADE with \texttt{MVT+MST} is in $O(DH + KDH +  K H )$.

\begin{table}[t]
\center
\renewcommand*{\arraystretch}{1.15}
\resizebox{.49\textwidth}{!}{
\setlength\tabcolsep{3.pt}
\begin{tabular}{r|c|c|c|c}
\toprule
\multicolumn{1}{c|}{\multirow{1}{*}{\bf Baselines}}    &    \multicolumn{4}{c}{\bf Features} \\ \cline{2-5}
\multicolumn{1}{c|}{\multirow{1}{*}{\bf (Related Works)}}		& \textit{NTM}		 & \textit{AuR}  & \textit{LVT} & \textit{GVT}$|$\textit{MVT}$|$\textit{MST} \\ \midrule
LDA  	      		&		 		&			&				  &		    			 \\ 
RSM         	&	\checkmark	&			&		  &		     			\\ 
DocNADE 		        &	\checkmark	&	\checkmark&		  &		     			\\ 
NVDM       			&	\checkmark	&			 &		  &		     			\\ 
ProdLDA    			 &		 		&			&		  &		    			\\ \hdashline

Gauss-LDA  	    				 &		 		&			&	\checkmark	  &		    			 \\ 		
glove-DMM 	 &		 		&			&	\checkmark	  &		    			\\ 
DocNADEe    	&	\checkmark	&	\checkmark&	\checkmark	  &		     			\\  \hdashline

EmbSum-Glove, EmbSum-BERT     								&		 		&			&		  &		     			\\ 
doc2vec   		&		 		&			&		  &		     			\\  \hdashline

{\bf this work}	   					&	\checkmark	 		&	\checkmark&	\checkmark	  &	\checkmark  \quad  \checkmark  \quad   \checkmark	\\ \bottomrule 
\end{tabular}}
\caption{Baselines (related works) vs this work. Here, {\it NTM} and {\it AuR} refer to neural network-based TM and autoregressive assumption, respectively.  
DocNADEe $\rightarrow$  DocNADE+Glove embeddings. 
}
\label{tab:baselinesvsthiswork}
\end{table}

\begin{table*}[t]
\center
\renewcommand*{\arraystretch}{1.15}
\resizebox{.85\textwidth}{!}{
\setlength\tabcolsep{3.pt}
\begin{tabular}{lr|c||ccc|ccc|ccc|ccc||cc}
\toprule
&  \multirow{1}{*}{\bf KBs from}   &     \multicolumn{1}{c||}{\bf Model} & \multicolumn{14}{c}{{\bf Scores on Target Corpus} ({\it in sparse-data  and sufficient-data settings})}   \\ \cline{4-17}

& \multirow{1}{*}{\bf Source}      &    \multicolumn{1}{c||}{\bf or Transfer} &   \multicolumn{3}{c|}{\texttt{20NSshort}}     &  \multicolumn{3}{c|}{\texttt{TMNtitle}}   &   \multicolumn{3}{c|}{\texttt{R21578title}}   &    \multicolumn{3}{c||}{\texttt{20NSsmall}}  &    \multicolumn{2}{c}{\texttt{TMN}}\\
&   \multirow{1}{*}{\bf Corpus}   &  \multicolumn{1}{c||}{\bf Type}  &   $PPL$  &  $COH$ &  $IR$ &    $PPL$  &  $COH$ &  $IR$ &  $PPL$  &  $COH$ &  $IR$ & $PPL$  &  $COH$ &  $IR$   & $PPL$  &  $COH$ \\ \midrule 

\multirow{3}{*}{\rotatebox{90}{\bf Baselines}}    &  \multirow{1}{*}{\it {\bf Baseline} TM}   &  NVDM    &  1047   & {.736}   & .076     &  973   & .740   & .190     &  372   & .735   & .271     &  957   & .515   & .090    &  833   & .673  \\ 
   &  \multirow{1}{*}{\it {\bf without} Word-} &  ProdLDA    &  923   & .689   & .062     &  1527   & .744   & .170     &  480   & {.742}   & .200     &  1181   & .394   & .062     &  1519   & .577  \\   
   &  \multirow{1}{*}{\it Embeddings}	    &   DocNADE    &  646   & .667   & .290     &  706   & .709   & .521     &  192   & .713   & .657     &  594   & .462   & .270   &  584   & .636 \\ \hline \hline  

\multirow{15}{*}{\rotatebox{90}{\bf Proposed}}    &  \multirow{3}{*}{\it 20NS}    &  LVT     &  {\bf 630}   & .673   & .298     &  705   & .709   & .523     &  194   & .708   & .656     &  594   & .455   & .288     		&  582   & .649 \\
					    &   &  GVT     &  646   & .690   & .303     &  718   & .720   & .527     &  {184}   & .698   & .660     &  594   & .500   & .310     &  590   & .652  \\
					     &  &  MVT &  {638}   & .690   & {\bf .314}     &  714   & .718   & .528     &  188   & .715   & .655     &  600   & .499   & .311    	 &  588   & .650   \\ \cline{2-17} 

   &  \multirow{3}{*}{\it TMN}    &  LVT     &  649   & .668   & .296     &  {\bf 655}   & .731   & .548     &  187   & .703   & .659     &  593   & .460   & .273    	&  -   & -   \\ 
					   &   &  GVT          &  661   & .692   & .294     &  689   & .728   & .555     &  191   & .709   & .660     &  596   & .521   & .276     	&  -   & -   \\ 
					   &   &  MVT   &  658   & .687   & .297     &  663   & .747   & .553     &  195   & .720   & .660     &  599   & .507   & .292     		&  -   & -   \\ \cline{2-17} 
					 
   &  \multirow{3}{*}{\it R21578}  &  LVT      &  656   & .667   & .292     &  704   & .715   & .522     &  186   & .715   & .676     &  593   & .458   & .267     	&  581   & .636  \\ 
					   &    &  GVT          &  654   & .672   & .293     &  716   & .719   & .526     &  194   & .706   & .672     &  595   & .485   & .279   	&  591   & .646   \\ 
					   &   &  MVT    &  650   & .670   & .296     &  716   & .720   & .528     &  194   & .724   & {.676}     &  599   & .490   & .280     	&  589   & .650  \\ \cline{2-17} 
					 
   &  \multirow{3}{*}{\it AGnews}  &  LVT      &  650   & .677   & .297     &  682   & .723   & .533     &  185   & .710   & .659     &  {\bf 592}   & .458   & .260     	&  {\bf 564}   & .668  \\ 
					    &   &  GVT          &  667   & .695   & .300     &  728   & .735   & .534     &  190   & .717   & .663     &  598   & .563   & .282   	&  601   & .684   \\ 
					    &  &  MVT    &  659   & .696   & .290     &  718   & .740   & .533     &  189   & .727   & .659     &  599   & .566   & .279     		&  592   & {.686}  \\  \cline{2-17} 
					 
   & \multirow{3}{*}{\it MST}  &  LVT      &  640   & .678   & .308     &  663   & .732   & .547     &  {\bf 182}   & .739   & .673     &  594   & .542   & .277   								&  568   	 & .674 \\


					   &    &  GVT          &  658   & .705   & .305     &  704   & .746   & .550     &  192   & .727   & .673     &  599   & .585   & {\bf .326} 						&  602   & .680  \\ 

	

				   & 	  &  MVT    &  656   & {\bf .740}   & {\bf .314}     &  680   & {\bf .752}   & {\bf .569}     &  188   & {\bf .745}   & {\bf .685}       &  600   & {\bf .637}   & .285    &  600   & {\bf .690}   \\ \hline \hline
		  

\multicolumn{3}{r||}{\bf Gain\%(vs DocNADE)} & {$\uparrow$2.48} & {$\uparrow$10.9} & {$\uparrow$8.28} & {$\uparrow$7.22} & {$\uparrow$6.06} & {$\uparrow$9.21} & {$\uparrow$5.20} & {$\uparrow$4.49} & {$\uparrow$4.26} & {$\uparrow$0.34} & {$\uparrow$37.9} & {$\uparrow$20.7} & {$\uparrow$3.42} & {$\uparrow$8.50} \\ \bottomrule

\end{tabular}}
\caption{State-of-the-art comparisons with TMs: Perplexity (PPL), topic coherence (COH) and precision@recall (IR) at retrieval fraction $0.02$. 
Scores reported on each of the target, given KBs from several sources.  
LVT and GVT employ {\small \texttt{WordPool}} and  {\small \texttt{TopicPool}}, respectively. MVT employs both. 
LVT+MST scores using {\it scheme (i)}. Here, Bold $\rightarrow$ Best score (in column) 
and Gain\% $\rightarrow$ Bold vs DocNADE.
}
\label{tab:scoresTMwithoutWordEmbeddings}
\end{table*}

\section{Evaluation and Analysis} 
{\bf Datasets}: Table \ref{tab:datadescription} describes 
the datasets used in high-resource source  
and low-and high-resource target domains for our experiments. 
The target domain ${\mathcal T}$ consists of four short-text corpora ({\texttt{20NSshort}}, {\texttt{TMNtitle}}, {\texttt{R21578title}} and {\texttt{Ohsumedtitle}}),
one small corpus ({\texttt{20NSsmall}}) and two large corpora ({\texttt{TMN}} and {\texttt{Ohsumed}}).  
However in source $\mathcal{S}$, we use five large corpora ({\texttt{20NS}}, {\texttt{R21578}}, {\texttt{TMN}}, {\texttt{AGnews}} and {\texttt{PubMed}}) 
in different label spaces (i.e, domains). Here, the corpora ($\mathcal{T}^5$, $\mathcal{T}^6$ and $\mathcal{S}^5$) belong to {\it medical} and others to {\it news}.

Additionally, Table \ref{tab:domainoverlap} suggests domain overlap (label match) in the target and source corpora, 
where we define 3 types of overlap: $\mathcal{I}$ (identical) if all labels match, $\mathcal{R}$ (related) if some labels match, and $\mathcal{D}$ (distant) if a very few or no labels match.
Note, our approaches are completely unsupervised and do not use the data labels ({\it appendix}). 

{\bf Reproducibility}:
We follow the experimental setup similar to DocNADE \citep{DBLP:conf/nips/LarochelleL12} and DocNADEe \citep{pankajgupta:2019iDocNADEe}, 
where the number of topics ($H$) is set to $200$. 
While DocNADEe requires the dimension (i.e., $E$) of word embeddings be the same as the latent topic (i.e., $H$), 
we follow {\it scheme (ii)} (Algorithm \ref{algo:computelogpv}) to introduce pre-trained word embeddings from 
Glove, FastText ($E$=300) \citep{DBLP:journals/tacl/BojanowskiGJM17} and BERT-base ($E$=768) models. 
See 
{\it appendix} 
for the experimental setup, hyperparameters 
and optimal values of $\lambda^k \in [0.1, 0.5, 1.0]$ and $\gamma^k \in [0.1, 0.01, 0.001]$. 

\begin{table*}[t]
\center
\renewcommand*{\arraystretch}{1.15}
\resizebox{.855\textwidth}{!}{
\setlength\tabcolsep{3.pt}
\begin{tabular}{lr|r||ccc|ccc|ccc|ccc||cc}
\toprule
& \multirow{1}{*}{\bf KBs from}   &     \multicolumn{1}{c||}{\bf Model} & \multicolumn{14}{c}{{\bf Scores on Target Corpus} ({\it in sparse-data  and sufficient-data settings})}   \\ \cline{4-17}

& \multirow{1}{*}{\bf Source}      &   \multicolumn{1}{c||}{\bf or Transfer} &   \multicolumn{3}{c|}{\texttt{20NSshort}}     &  \multicolumn{3}{c|}{\texttt{TMNtitle}}   &   \multicolumn{3}{c|}{\texttt{R21578title}}   &    \multicolumn{3}{c||}{\texttt{20NSsmall}}  &    \multicolumn{2}{c}{\texttt{TMN}}\\
&    \multirow{1}{*}{\bf Corpus}   &   \multicolumn{1}{c||}{\bf Type}  &   $PPL$  &  $COH$ &  $IR$ &    $PPL$  &  $COH$ &  $IR$ &  $PPL$  &  $COH$ &  $IR$ & $PPL$  &  $COH$ &  $IR$   & $PPL$  &  $COH$ \\ \midrule 

\multirow{7}{*}{\rotatebox{90}{\bf Baselines}}  &	&  doc2vec        			&  -   & -       & .090     &  -   & -        & .190      &  -   & -        & .518     &  -   & -   & .200    &  -   & -   \\ 
&      							&  EmbSum-Glove   		&  -   & -   & .236     &  -   &  -   & .513     &  -   &  -   & .587     &  -   & -   & .214     &  -   & -   \\   
&							&  EmbSum-BERT    	&  -   & -   & .261     &  -   &  -   & .499     &  -   &  -   & .594     &  -   & -   & .262     &  -   & -   \\   \cdashline{2-17}  
&	\multirow{1}{*}{{\bf Baseline} TM}    &  Gauss-LDA    &  -   & -   & .080     &  -   & -   & .408     &  -   & -   & .367     &  -   & -   & .090     &  -   & - \\ 
&	\multirow{1}{*}{\it {\bf with} Word-} 	      &  glove-DMM    &  -   & .512   & .183     &  -   & .633   & .445     &  -   & .364   & .273     &  -   & .578   & .090   &  -   & .705   \\ 
& 	\multirow{1}{*}{\it Embeddings}		&  $\rightarrow$ DocNADEe    &  {\bf 629}   & .674   & .294     &  680   & .719   & .540     &  187   & .721   & .663     &  {\bf 590}   & .455   & .274   &  {572}  & .664 \\ \hline\hline 


                                          
\multirow{8}{*}{\rotatebox{90}{\bf Proposed}}  & 	\multirow{1}{*}{\it 20NS}    &  MVT+Glove    &  \underline{630}   & .721   & {.320}     &  688   & .741   & .565     &  {\bf 183}   & .724   & .667     &  597   & .561   & {\bf .306}    &  {\bf 570}   & .693  \\  

                                        

& 	\multirow{1}{*}{\it TMN}   &  MVT+Glove     &  640   & .731   & .295     &  {\bf 673}   & .750   & {.576}     &  {184}   & .716   & .672     &  599   & .594   & .261    	 &  -   & -  \\  

                                              

& 	\multirow{1}{*}{\it R21578}     &  MVT+Glove     &  633   & .705   & .295     &  689   & .738   & .540     &  185   & { .737}   & {\bf .691}     &  \underline{595}   & .485   & .255    	&  577   & .697  \\  

                                            

& 	\multirow{1}{*}{\it AGnews}   &  MVT+Glove    &  642   & .734   & .302     &  706   & .748   & .565     &  190   & { .734}   & .675     &  598   & .573   & .284     	&  585   & {.703}   \\  \cline{2-17} 



& 	\multirow{3}{*}{\it MST}   &  MVT+Glove     &  644   & {.739}   & .304     &  {\bf 673}   & {\bf .752}   & .570     &  {\bf 183}   & {.742}   & {.684}     &  598   & {.631}   & {.282}   &  582   & {.710}   \\ 
& 								&     + FastText    &  654   & {.741}   & .313     & {\bf  673}   & .751   & {.578}     &  {\bf 183}   & {.744}   & .684     &  599   & {.634}   & .254    	&  582   & {\bf .711}  \\  
& 								&  	 + BERT       &  -   & {\bf .744}   & {\bf .322}     &  {-}   & {\bf .752}   & {\bf .604}     &  {-}   & {\bf .745}   & {.680}     &  -   & {\bf .640}   & {.282}   &  -   & {.709}   \\ 
\hline\hline
\multicolumn{3}{r||}{\bf Gain\% (vs DocNADEe)} & {$\downarrow$0.16} & {$\uparrow$10.4} & {$\uparrow$9.5} & {$\uparrow$1.03} & {$\uparrow$4.60} & {$\uparrow$11.9} & {$\uparrow$3.33} & {$\uparrow$3.20} & {$\uparrow$4.22} & {$\downarrow$0.85} & {$\uparrow$40.7} & {$\uparrow$2.92} & {$\uparrow$.35} & {$\uparrow$7.08} \\ \bottomrule
\end{tabular}}
\caption{State-of-the-art comparisons against baseline TMs using context-insensitive word embeddings: 
PPL,  COH and IR at retrieval fraction $0.02$. 
Scores are reported on each of the target, given the KBs. 
Here, MVT $\rightarrow$ LVT+GVT, DocNADEe $\rightarrow$ DocNADE+Glove, Bold $\rightarrow$ Best score (in column), Underline $\rightarrow$ Second best score (in column) and 
Gain\% $\rightarrow$ Bold vs DocNADEe. For all the configurations, we apply a projection on ([non-]contextualized) word embeddings from several sources, i.e., {\it scheme (ii)}. 
}
\label{tab:scoresTMwithWordEmbeddings}
\end{table*}

{\bf Baselines (Related Works)}: 
(1) {\it Topic Models without Transfer Learning} that learn topics in isolation using the given target corpus only. 
We employ LDA-based variant, i.e., ProdLDA \cite{SrivastavaSutton} and neural network-based variants, i.e., 
DocNADE (autoregressive) 
and NVDM (non-autoregressive) \cite{DBLP:conf/icml/MiaoYB16}. 

(2) {\it Topic Models with Transfer Learning} that 
leverages 
pre-trained context-insensitive word embeddings \cite{D14-1162}.  
We consider topic models based on both LDA, i.e., Gauss-LDA \cite{P15-1077} and glove-GMM \cite{DBLP:journals/tacl/NguyenBDJ15}, 
and neural networks, i.e., DocNADEe \cite{pankajgupta:2019iDocNADEe}. 
They do not leverage pre-trained topic-embeddings (i.e., {\it GVT}), contextualized word-embedding and MST-MVT techniques. 

(3) {\it Unsupervised Document Representation} to quantify the quality of document representations. 
We use 3 strategies: doc2vec \cite{DBLP:conf/icml/LeM14}, EmbSum-Glove and EmbSum-BERT 
(represent a document by summing the pre-trained embeddings of it’s words from Glove and BERT).

(4) {\it Zero-shot Topic Modeling} to demonstrate transfer learning capabilities of the proposed framework, 
where we build (train) a TM using all source corpora and evaluate on the target corpus $\mathcal{T}$, and 

(5) {\it Data-augmentation} that first augments the target corpus with all the source corpora and then
builds a TM 
to evaluate transfer learning on $\mathcal{T}$. 

Table~\ref{tab:baselinesvsthiswork} summarizes the comparison of this work with the aforementioned baselines. Tables \ref{tab:scoresTMwithoutWordEmbeddings} and \ref{tab:scoresTMwithWordEmbeddings} employ baseline TMs without and with transfer learning, respectively.

\makeatletter
\def\labelonly{BDF}
\def\labelcheck#1{
    \edef\pgfmarshal{\noexpand\pgfutil@in@{#1}{\labelonly}}
    \pgfmarshal
    \ifpgfutil@in@[#1]\fi
}
\makeatother
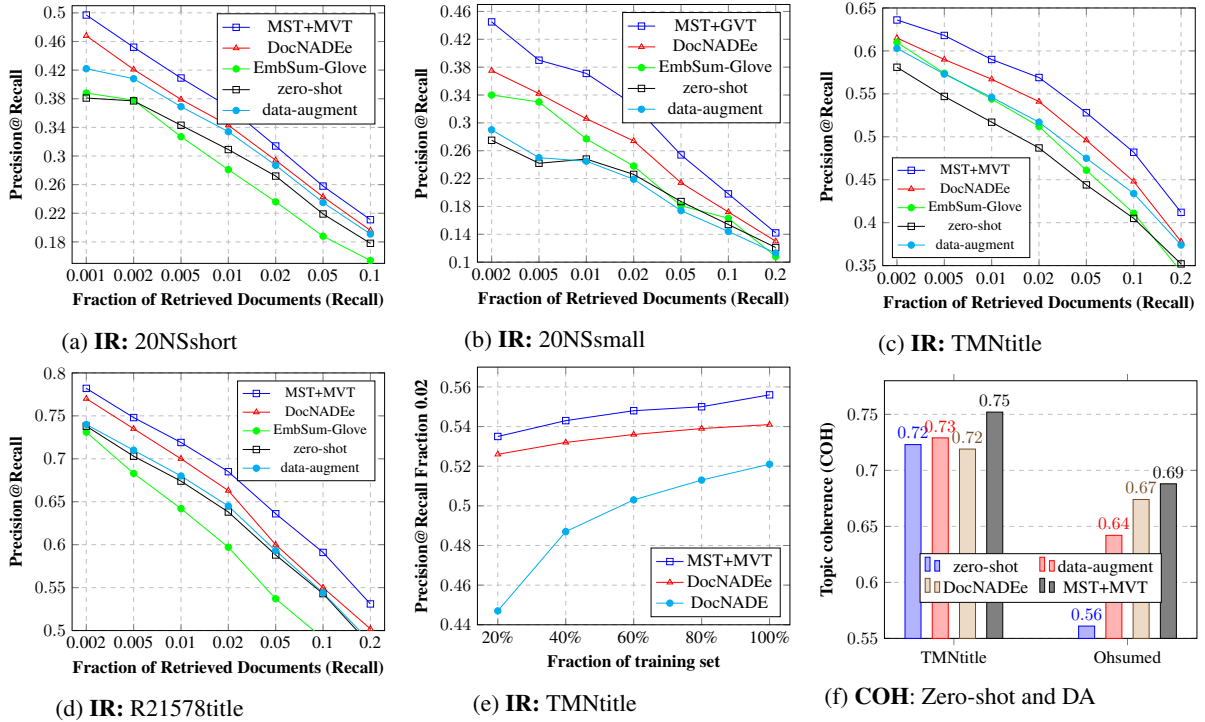
\begin{figure*}[t]
\center
\begin{subfigure}{0.24\textwidth}
\center
\begin{tikzpicture}[scale=0.6][baseline]
\begin{axis}[
    xlabel={\bf Fraction of Retrieved Documents (Recall)},
    ylabel={\bf Precision@Recall}, 
    xmin=-0.3, xmax=6.3,
    ymin=0.15, ymax=0.51,
   /pgfplots/ytick={.10,.14,...,.51},
    xtick={0,1,2,3,4,5,6},
    xticklabels={0.001, 0.002, 0.005, 0.01, 0.02, 0.05, 0.1}, 
    legend pos=north east,
    ymajorgrids=true, xmajorgrids=true,
    grid style=dashed,
    legend style={font=\normalsize}
]
\addplot[
	color=blue,
	mark=square,
	]
	plot coordinates {
    (0, 0.497)
    (1, 0.452)
    (2, 0.409)
    (3, 0.369)
    (4, 0.314)
    (5, 0.258)
    (6, 0.211)
	};
\addlegendentry{MST+MVT}

\addplot[
	color=red,
	mark=triangle,
	]
	plot coordinates {
    (0, 0.468)
    (1, 0.421)
    (2, 0.379)
    (3, 0.343)
    (4, 0.294)
    (5, 0.243)
    (6, 0.196)
	};
\addlegendentry{DocNADEe}

\addplot[
	color=green,
	mark=*,
	]
	plot coordinates {
    (0, 0.388)
    (1, 0.378)
    (2, 0.327)
    (3, 0.281)
    (4, 0.236)
    (5, 0.188)
    (6, 0.154)
	};
\addlegendentry{EmbSum-Glove}

\addplot[
	color=black,
	mark=square,
	]
	plot coordinates {
    (0, 0.381)
    (1, 0.377)
    (2, 0.343)
    (3, 0.309)
    (4, 0.272)
    (5, 0.219)
    (6, 0.178)
	};
\addlegendentry{zero-shot}

\addplot[
	color=cyan,
	mark=*,
	]
	plot coordinates {
    (0, 0.422)
    (1, 0.408)
    (2, 0.369)
    (3, 0.334)
    (4, 0.287)
    (5, 0.235)
    (6, 0.191)
	};
\addlegendentry{data-augment}
\end{axis}
\end{tikzpicture}%
\caption{{\bf IR:} 20NSshort} \label{IR20NSshort}
\end{subfigure}\hspace*{\fill}%
\begin{subfigure}{0.24\textwidth}
\centering
\begin{tikzpicture}[scale=0.6][baseline]
\begin{axis}[
    xlabel={\bf Fraction of Retrieved Documents (Recall)},
    ylabel={\bf Precision@Recall},
    xmin=-0.3, xmax=6.3,
    ymin=0.10, ymax=0.47,
   /pgfplots/ytick={.10,.14,...,.47},
    xtick={0,1,2,3,4,5,6},
    xticklabels={0.002, 0.005, 0.01, 0.02, 0.05, 0.1,  0.2},
    legend pos=north east,
    ymajorgrids=true, xmajorgrids=true,
    grid style=dashed,
     legend style={font=\normalsize}
]

\addplot[
	color=blue,
	mark=square,
	]
	plot coordinates {
    (0, 0.445)
    (1, 0.390)
    (2, 0.371)
    (3, 0.326)
    (4, 0.254)
    (5, 0.198)
    (6, 0.142)
	};
\addlegendentry{MST+GVT}

\addplot[
	color=red,
	mark=triangle,
	]
	plot coordinates {
    (0, 0.375)
    (1, 0.342)
    (2, 0.306)
    (3, 0.274)
    (4, 0.214)
    (5, 0.172)
    (6, 0.130)
	};
\addlegendentry{DocNADEe}

\addplot[
	color=green,
	mark=*,
	]
	plot coordinates {
    (0, 0.340)
    (1, 0.330)
    (2, 0.277)
    (3, 0.238)
    (4, 0.181)
    (5, 0.163)
    (6, 0.108)
	};
\addlegendentry{EmbSum-Glove}

\addplot[
	color=black,
	mark=square,
	]
	plot coordinates {
    (0, 0.275)
    (1, 0.242)
    (2, 0.248)
    (3, 0.226)
    (4, 0.187)
    (5, 0.154)
    (6, 0.121)
	};
\addlegendentry{zero-shot}

\addplot[
	color=cyan,
	mark=*,
	]
	plot coordinates {
    (0, 0.290)
    (1, 0.250)
    (2, 0.245)
    (3, 0.219)
    (4, .174)
    (5, 0.144)
    (6, 0.113)
	};
\addlegendentry{data-augment}
\end{axis}
\end{tikzpicture}%
\caption{{\bf IR:} 20NSsmall} \label{IR20NSsmall}
\end{subfigure}\hspace*{\fill}%
\begin{subfigure}{0.24\textwidth}
\centering
\begin{tikzpicture}[scale=0.6][baseline]
\begin{axis}[
    xlabel={\bf Fraction of Retrieved Documents (Recall)},
    ylabel={\bf Precision@Recall},
    xmin=-0.3, xmax=6.3,
    ymin=0.35, ymax=0.65,
   /pgfplots/ytick={.35,.40,...,.65},
    xtick={0,1,2,3,4,5,6},
    xticklabels={0.002, 0.005, 0.01, 0.02, 0.05, 0.1,  0.2},
    legend pos=south west,
     legend style={font=\small},
    ymajorgrids=true, xmajorgrids=true,
    grid style=dashed,
    legend style={font=\small}
]

\addplot[
	color=blue,
	mark=square,
	]
	plot coordinates {
    (0, 0.636)
    (1, 0.618)
    (2, 0.590)
    (3, 0.569)
    (4, 0.528)
    (5, 0.482)
    (6, 0.412)
	};
\addlegendentry{MST+MVT}

\addplot[
	color=red,
	mark=triangle,
	]
	plot coordinates {
    (0, 0.615)
    (1, 0.590)
    (2, 0.567)
    (3, 0.541)
    (4, 0.496)
    (5, 0.448)
    (6, 0.378)
	};
\addlegendentry{DocNADEe}

\addplot[
	color=green,
	mark=*,
	]
	plot coordinates {
    (0, 0.610)
    (1, 0.574)
    (2, 0.544)
    (3, 0.512)
    (4, 0.461)
    (5, 0.411)
    (6, 0.343)
    (7, 0.296)
    (8, 0.239)
    (9, 0.190)
    (10, 0.169)
	};
\addlegendentry{EmbSum-Glove}

\addplot[
	color=black,
	mark=square,
	]
	plot coordinates {
    (0, 0.581)
    (1, 0.547)
    (2, 0.517)
    (3, 0.487)
    (4, 0.444)
    (5, 0.405)
    (6, 0.352)
	};
\addlegendentry{zero-shot}

\addplot[
	color=cyan,
	mark=*,
	]
	plot coordinates {
    (0, 0.603)
    (1, 0.573)
    (2, 0.546)
    (3, 0.517)
    (4, 0.475)
    (5, 0.434)
    (6, 0.374)
	};
\addlegendentry{data-augment}
\end{axis}
\end{tikzpicture}%
\caption{{\bf IR:} TMNtitle} \label{IRTMNtitle}
\end{subfigure}\hspace*{\fill}%

\begin{subfigure}{0.24\textwidth}
\centering
\begin{tikzpicture}[scale=0.6][baseline]
\begin{axis}[
    xlabel={\bf Fraction of Retrieved Documents (Recall)},
    ylabel={\bf Precision@Recall},
    xmin=-0.3, xmax=6.3,
    ymin=0.50, ymax=0.80,
   /pgfplots/ytick={.5,.55,...,.80},
    xtick={0,1,2,3,4,5,6},
    xticklabels={0.002, 0.005, 0.01, 0.02, 0.05, 0.1,  0.2},
    legend pos=north east,
    ymajorgrids=true, xmajorgrids=true,
    grid style=dashed,
    legend style={font=\small}
]

\addplot[
	color=blue,
	mark=square,
	]
	plot coordinates {
    (0, 0.782)
    (1, 0.748)
    (2, 0.719)
    (3, 0.685)
    (4, 0.636)
    (5, 0.591)
    (6, 0.531)
	};
\addlegendentry{MST+MVT}

\addplot[
	color=red,
	mark=triangle,
	]
	plot coordinates {
    (0, 0.77)
    (1, 0.735)
    (2, 0.70)
    (3, 0.663)
    (4, 0.60)
    (5, 0.55)
    (6, 0.502)
	};
\addlegendentry{DocNADEe}

\addplot[
	color=green,
	mark=*,
	]
	plot coordinates {
    (0, 0.731)
    (1, 0.683)
    (2, 0.642)
    (3, 0.597)
    (4, 0.537)
    (5, 0.491)
    (6, 0.442)
	};
\addlegendentry{EmbSum-Glove}

\addplot[
	color=black,
	mark=square,
	]
	plot coordinates {
    (0, 0.738)
    (1, 0.703)
    (2, 0.674)
    (3, 0.638)
    (4, 0.588)
    (5, 0.543)
    (6, 0.483)
	};
\addlegendentry{zero-shot}

\addplot[
	color=cyan,
	mark=*,
	]
	plot coordinates {
    (0, 0.740)
    (1, 0.710)
    (2, 0.680)
    (3, 0.645)
    (4, 0.593)
    (5, 0.544)
    (6, 0.487)
	};
\addlegendentry{data-augment}
\end{axis}
\end{tikzpicture}%
\caption{{\bf IR:} R21578title} \label{IRR21578title}
\end{subfigure}\hspace*{\fill}%
\begin{subfigure}{0.24\textwidth}
\centering
\begin{tikzpicture}[scale=0.6][baseline]
\begin{axis}[
    xlabel={\bf Fraction of training set},
    ylabel={\bf Precision@Recall Fraction 0.02},
     xmin=-0.3, xmax=4.3,
    ymin=0.44, ymax=0.57,
   /pgfplots/ytick={.44,.46,...,.57},
    xtick={0, 1, 2, 3, 4},
    xticklabels={20\%, 40\%, 60\%, 80\%, 100\%},
    legend pos=south east,
    ymajorgrids=true, xmajorgrids=true,
    grid style=dashed,
     legend style={font=\normalsize}
]
\addplot[
	color=blue,
	mark=square,
	]
	plot coordinates {
     (0, 0.535)
    (1, 0.543)
    (2, 0.548)
    (3, 0.550)
    (4, 0.556) 
	};
\addlegendentry{MST+MVT}

\addplot[
	color=red,
	mark=triangle,
	]
	plot coordinates {
    (0, 0.526)
    (1, 0.532)
    (2, 0.536)
    (3, 0.539)
    (4, 0.541) 
	};
\addlegendentry{DocNADEe}

\addplot[
	color=cyan,
	mark=*,
	]
	plot coordinates {
    (0, 0.447)
    (1, 0.487)
    (2, 0.503)
    (3, 0.513)
    (4, 0.521) 
	};
\addlegendentry{DocNADE}
\end{axis}
\end{tikzpicture}%
\caption{{\bf IR:} TMNtitle} \label{TMNtitle}
\end{subfigure}\hspace*{\fill}%
\begin{subfigure}{0.24\textwidth}
\centering
\begin{tikzpicture}[scale=0.6][baseline]
\begin{axis}[
    legend pos=north east,
    legend columns=2, 
        legend style={
            /tikz/column 2/.style={
                column sep=5pt,
            },
        },
    ybar=7pt,
    legend style={at={(0.5,0.33)},
    anchor=north,legend columns=2},
    xmin=-0.4, xmax=1.4,
    ymin=0.55, ymax=.780,
    ylabel={\bf Topic coherence (COH)},
    xtick={0, 1},
    xticklabels={TMNtitle, Ohsumed},
    nodes near coords,
    nodes near coords align={vertical},
    ymajorgrids=true, 
    grid style=dashed,
    ]
\addplot coordinates {(0,.723) (1,.561)}; \addlegendentry{zero-shot}
\addplot coordinates {(0,.729) (1,.642)}; \addlegendentry{data-augment}
\addplot coordinates {(0,.719) (1,.674)}; \addlegendentry{DocNADEe}
\addplot coordinates {(0,.752) (1,.688)}; \addlegendentry{MST+MVT}
\end{axis}
\end{tikzpicture}
\caption{{\bf COH}: Zero-shot and DA} \label{COH:ZeroshotDataAugTMNtitle}
\end{subfigure}\hspace*{\fill}%
\caption{
(a, b, c, d) Retrieval performance (precision@recall) on 4 datasets: 20NSshort, 20NSsmall, TMNtitle and R21578title. 
(e) Precision at recall fraction $0.02$, each for a fraction (20\%, 40\%, 60\%, 80\%, 100\%) of the training set of TMNtitle. 
(f) Zero-shot and data-augmentation (DA) for COH on TMNtitle and Ohsumed.}
\label{fig:docretrieval}
\end{figure*}

\subsection{Generalization: Perplexity (PPL)} 
To evaluate generative performance of DocNADE-based NTM, 
we compute average held-out perplexity per word: 
{\small $PPL$ = $\exp \big( - \frac{1}{N} \sum_{t=1}^{N} \frac{1}{|{\bf v}_t|} \log p({\bf v}_{t}) \big)$}, 
where $N$  and $|{\bf v}_t|$ are the number of documents and words in a document ${\bf v}_{t}$, respectively.  

Tables \ref{tab:scoresTMwithoutWordEmbeddings} and \ref{tab:scoresTMwithWordEmbeddings} quantitatively show PPL scores on the five target corpora 
using one or four sources. 
In Table \ref{tab:scoresTMwithoutWordEmbeddings} using  {\texttt{TMN}} (as a single source) for LVT, GVT and MVT transfer types on the target {\texttt{TMNtitle}}, 
we see improved (reduced) PPL scores: ($655$ vs $706$), ($689$ vs $706$)  and ($663$ vs $706$) respectively in comparison to DocNADE. 
We also observe gains due to MST+LVT, MST+GVT and MST+MVT configurations on {\texttt{TMNtitle}}.  
Similarly in MST+LVT for {\texttt{R21578title}}, we observe a gain of 5.2\% (182 vs 192), suggesting that 
multi-source transfer learning using pretrained word and topic embeddings (jointly) helps improving TM, 
and it also verifies domain relatedness (e.g., in {\texttt{TMN}}-{\texttt{TMNtitle}} and {\texttt{AGnews}}-{\texttt{TMN}}).     
Similarly, Table \ref{tab:scoresTMwithWordEmbeddings} reports gains in PPL (e.g., on {\texttt{TMNtitle}}, {\texttt{R21578title}}, etc.) compared to the baseline DocNADEe. 
PPL scores due to BERT can be not computed since its embeddings are aware of both preceding and following contexts. 

In Table \ref{tab:scoresMedicaldomain}, we show PPL scores on 2 medical target corpora: {\texttt{Ohsumtitle}}  and {\texttt{Ohsumed}} 
using 2 sources: {\texttt{AGnews}}  ({\it news}) and {\texttt{PubMed}} ({\it medical}) to perform cross-domain and in-domain transfers. 
We see that using {\texttt{PubMed}} for LVT on both the targets improves generalization. 
Overall, we report a gain of 17.3\% ($1268$ vs $1534$) on {\texttt{Ohsumedtitle}} 
and  8.55\% ($1497$ vs $1637$) on {\texttt{Ohsumed}} datasets, compared to DocNADEe. 

\subsection{Interpretabilty: Topic Coherence (COH)}
While PPL is used for model selection, 
\citet{DBLP:conf/nips/ChangBGWB09} showed in some cases humans preferred TMs (based on the semantic quality of topics) 
with higher (worse) perplexities. 
Therefore, we also estimate the quality of topics. 
We follow \citet{DBLP:conf/wsdm/RoderBH15} and \citet{pankajgupta:2019iDocNADEe} to compute COH of the top 10 words in each topic. 
Essentially, the higher scores imply the coherent topics. 

Tables \ref{tab:scoresTMwithoutWordEmbeddings} and \ref{tab:scoresTMwithWordEmbeddings} (under {COH} column) demonstrate  that 
our approaches (GVT, MVT and MST) 
show noticeable gains 
and thus improve topic quality. For instance in Table \ref{tab:scoresTMwithoutWordEmbeddings},  
when {\texttt{AGnews}} is used as a single source for {\texttt{20NSsmall}} datatset,  
we observe a gain in COH due to GVT (.563 vs .462) and MVT (.566 vs .462). Additionally, noticeable gains are reported due to 
MST+LVT (.542 vs .462),  MST+GVT (.585 vs .462) and MST+MVT (.637 vs .462), compared to DocNADE. 
Importantly, we find a trend MVT$>$GVT$>$LVT in COH scores for both the single-source and multi-source transfers. 
Similarly, Table \ref{tab:scoresTMwithWordEmbeddings} show noticeable gains (e.g., 40.7\%, 10.4\%, 7.08\%, etc.) in COH due to MST+MVT+Glove +FastText+BERT setting. 
Moreover, Table \ref{tab:scoresMedicaldomain} shows gains in COH due to GVT on {\texttt{Ohsumedtitle}} and   {\texttt{Ohsumed}},  using pretrained knowledge from {\texttt{PubMed}}.
Overall, the GVT, MVT and MST boost {COH} for all the five target corpora compared to the baseline TMs (i.e., DocNADE and DocNADEe). 
The improvements suggest that the approaches scale across domains.

\begin{table}[t]
\center
\renewcommand*{\arraystretch}{1.15}
\resizebox{.48\textwidth}{!}{
\setlength\tabcolsep{3.pt}
\begin{tabular}{r|r||ccc|ccc}
\multirow{1}{*}{\bf KBs from}   &      \multicolumn{1}{c||}{\bf Model} & \multicolumn{6}{c}{{\bf Scores on Target Corpus}}   \\ \cline{3-8}

\multirow{1}{*}{\bf Source}      &   \multicolumn{1}{c||}{\bf or Transfer} &   \multicolumn{3}{c|}{\texttt{Ohsumedtitle}}     &  \multicolumn{3}{c}{\texttt{Ohsumed}}   \\
   \multirow{1}{*}{\bf Corpus}   &    \multicolumn{1}{c||}{\bf Type} & $PPL$  &  $COH$ &  $IR$ &    $PPL$  &  $COH$ &  $IR$   \\ \midrule 

\multirow{4}{*}{\bf baselines}  &  ProdLDA    	&  1121   & .734   & .080     &  1677   & .646   & .080      \\ 
						&   DocNADE     &  1321   & .728   & .160     &  1706   & .662   & .184          \\ 

 					       &  EmbSum-BioEmb       &  -   & -   & .150     &  -   & -   & .148          \\   
						
						&  EmbSum-SciBERT       &  -   & -   & .160     &  -   & -   & .165          \\   

						&  DocNADEe    &   1534   & .738   & .175     &  1637   & .674   &  .183          \\   \hline \hline

\multirow{4}{*}{\it AGnews}    &  LVT     		 &  1587   & .732   & .160    &  1717   & .657   & .184          \\
					  &  GVT     		&  1529   & .732   & .160     &  1594   & .665   & .185          \\
					  &  MVT   			 &  1528   & .734   & .160    &  1598   & .666   & .184          \\
 					&  + BioEmb   		&  1488   & .747   & .176     &  1595   & .681   & .187          \\ \hline

\multirow{4}{*}{\it PubMed}    &  LVT     		 &  {\bf 1268}   & .732   & .172     &  1535   & .669   & .190          \\
					  &  GVT       		&  1392   & .740   & .173     	&  1718   & .671   & {\bf .192}          \\
					  &  MVT    		&  1408   & .743   & .178     	&  1514   & .674   & .191          \\

 					&  + BioEmb     		&  1364   & {\bf .753}   & {.182}     	&  1633   & {\bf .689}   & .191         \\ \hline \hline
\multirow{4}{*}{\it MST}  &  LVT      			&  {\bf 1268}   & .733   & .172     &  1536   & .668   & .190          \\  
					  &  GVT       		&  1391   & .740   & .172     	&  1504   & .666   & {\bf .192}        \\
 					&  MVT    		&  1399   & .744   & .177     	&  1607   & .679   & .191          \\
					 &  + BioEmb    		 &  1375   & .751   & .180    	 &  {\bf 1497}   & .693   & .190        \\ 
					&  + BioFastText    		 &  1350   & {\bf .753}   & .178    	 &  1641   & .688   & .187       \\
					&  + SciBERT    		 &  -   & {\bf .753}   & {\bf .183}    	 &  -   & .682   & .182   \\ \hline \hline                   

\multicolumn{2}{r||}{\bf Gain\% (vs DocNADE)} & {$\uparrow$4.01} & {$\uparrow$3.43} & {$\uparrow$14.4} & {$\uparrow$12.3} & {$\uparrow$4.08} & {$\uparrow$4.35} \\
\multicolumn{2}{r||}{\bf Gain\% (vs DocNADEe)} & {$\uparrow$17.3} & {$\uparrow$2.03} & {$\uparrow$4.60} & {$\uparrow$8.55} & {$\uparrow$2.22} & {$\uparrow$4.91} 
\end{tabular}}
\caption{
PPL, COH, and IR at fraction $0.02$.  
BioEmb and BioFastText \cite{moen2013distributional}: 200-dimension; 
SciBERT: Pretrained BERT-variant \cite{scibert2019}. 
+ BioEmb: MVT+BioEmb
}
\label{tab:scoresMedicaldomain}
\end{table}

\subsection{Applicability: Information Retrieval (IR)}

We further evaluate the quality of document representations and perform an IR task 
using the  label information only to compute precision.  
We follow the experimental setup similar to \citet{pankajgupta:2019iDocNADEe}.
See the details in 
{\it appendix}. 

Tables \ref{tab:scoresTMwithoutWordEmbeddings} and \ref{tab:scoresTMwithWordEmbeddings} report precision scores 
at retrieval fraction $0.02$ 
where the configuration MST+MVT outperforms both the DocNADE and DocNADEe for all 4 targets. 
We observe {\it large gains} in precision: 
(a) Table \ref{tab:scoresTMwithoutWordEmbeddings}: 20.7\% (.326 vs .270) on  {\texttt{20NSsmall}}, 9.21\% (.569 vs .521) on  {\texttt{TMNtitle}}, etc.,
(b) Table \ref{tab:scoresTMwithWordEmbeddings}: 11.9\% (.604 vs .540) on  {\texttt{TMNtitle}} and 9.5\% (.322 vs .294) on  {\texttt{20NSshort}}, etc.,
(c)  Table \ref{tab:scoresMedicaldomain}: 14.4\% (.183 vs .160) on {\texttt{Ohsumedtitle}}. 
Additionally, Figures \ref{IR20NSshort}, \ref{IR20NSsmall}, \ref{IRTMNtitle} and \ref{IRR21578title} illustrate precision-recall curves on  
{\texttt{20NSshort}}, {\texttt{20NSsmall}}, {\texttt{TMNtitle}} and {\texttt{R21578title}} respectively, 
where MST+MVT and MST+GVT consistently outperform the baselines at all fractions. 



\subsection{Zero/Few-shot and Data-augmentation} 
Figures \ref{IR20NSshort}, \ref{IR20NSsmall}, \ref{IRTMNtitle} and \ref{IRR21578title} 
show precision in the {\it zero-shot} (source-only training) and {\it data-augmentation} 
(source+target training) configurations. 
Observe that the latter helps in learning meaningful representations and performs better than zero-shot; however, 
it is outperformed by MST+MVT, suggesting that a naive (data space) augmentation does not add sufficient prior or relevant information to the sparse target.    
Thus, we find that it is beneficial to augment training data in feature space (e.g., LVT, GVT and MVT) 
especially for unsupervised topic models using \texttt{WordPool} and \texttt{TopicPool}. 

Moreover in the {\it few-shot} setting, we first split the training data of {\texttt{TMNtitle}} into several sets: 20\%, 40\%, 60\%, 80\% of the training set 
and then retrain DocNADE, DocNADEe and DocNADE+MST+MVT on each as a sparse target. 
We demonstrate transfer learning in such sparse-data settings using the KBs: \texttt{WordPool} and \texttt{TopicPool} jointly. 
Figure \ref{TMNtitle} plots precision at retrieval fraction $0.02$ and 
validates that the proposed modeling consistently outperforms both the baselines: DocNADE and DocNADEe. 

Beyond IR, we further investigate computing topic coherence (COH) for the zero-shot and data-augmentation baselines, where 
the COH scores in Figure \ref{COH:ZeroshotDataAugTMNtitle} suggest that MST+MVT outperforms DocNADEe, zero-shot and data-augmentation. 


\begin{table}[t]
      \center
	\renewcommand*{\arraystretch}{1.1}
	\resizebox{.48\textwidth}{!}{
\begin{tabular}{c|c|c|l}
{\bf $\mathcal{T}$}   &     {\bf $\mathcal{S}$}        & \multicolumn{1}{c|}{\bf Model}      & \multicolumn{1}{c}{\bf Topic-words (Top 5)}     \\ \hline
\multirow{6}{*}{\rotatebox{90}{\texttt{20NSshort}}}          &	\multirow{1}{*}{\texttt{20NS}}   	       &  DocNADE			& {\color{blue} shipping}, sale, prices, {\color{blue} expensive}, price		\\  \cdashline{2-4}  
 							      &			       					    &  {\bf -}GVT			   &	sale, price, monitor, {\color{red} site}, {\color{red} setup}		    \\    
 						             &		   							    &     {\bf +}GVT				&      {\color{blue} shipping}, sale, price, {\color{blue} expensive}, subscribe			\\     \cline{2-4}       

         &	 \multirow{1}{*}{\texttt{AGnews}} 	         &  DocNADE			&	{\color{blue} microsoft}, {\color{blue} software}, ibm, linux, {\color{blue} computer}		 \\    \cdashline{2-4}
							         &								     	& {\bf -}GVT			&	apple, modem, {\color{red} side}, baud, {\color{red} perform}		   \\    
 						                &					  				&  	{\bf +}GVT		&      {\color{blue} microsoft}, {\color{blue} software}, desktop, {\color{blue} computer}, apple \\	\hline

\multirow{4}{*}{\rotatebox{90}{\texttt{TMNtitle}}}         &		\multirow{1}{*}{\texttt{AGnews}}  	 &  DocNADE 				&	miners,  earthquake, {\color{blue} explosion},  stormed, quake 		\\   \cdashline{2-4}
						          &		\multirow{1}{*}{\texttt{TMN}}  		 &  DocNADE 				&	tsunami, quake, japan, earthquake, {\color{blue} radiation} 		\\   \cdashline{2-4}
						           &									        &  {\bf -}GVT			&	{\color{red} strike}, {\color{red} jackson}, kill, earthquake, injures		  \\    
 						             &									&  	{\bf +}GVT					&      earthquake, {\color{blue} radiation}, {\color{blue} explosion}, wildfire  		 
\end{tabular}}
\caption{Source $\mathcal{S}$ and target $\mathcal{T}$ topics before (-) and after (+) topic transfer (GVT) from one/more source(s)}
\label{topiccoherenceexamples}
\end{table}

\begin{table}[t]
\centering
	\renewcommand*{\arraystretch}{1.1}
	\resizebox{.47\textwidth}{!}{
\begin{tabular}{ccc|c|c} 
\multicolumn{3}{c|}{source corpora} &  \multicolumn{2}{c}{target corpus}         \\ \hline
\multirow{2}{*}{\texttt{20NS}}	&	\multirow{2}{*}{\texttt{R21578}}	&  \multirow{2}{*}{\texttt{AGnews}}		& \multicolumn{2}{c}{\texttt{20NSshort}} \\
	               &						&					&	-GVT	 	&	+GVT  \\  \hline 
key			&		chips 			&	chips 			&	virus	 &	chips 	 		\\
encrypted		&		semiconductor 		&	chipmaker 			&	intel	  &	technology	 \\  
{\color{blue} encryption}		&		miti 				&	 processors 		&	{\color{red} gosh}	 &	intel		 \\  
{\color{blue} clipper}		&		makers 			&	 semiconductor 		&	crash	 &	{\color{blue} encryption}				 \\  
keys			&		semiconductors 		&	intel 				&	chips	   &	{\color{blue} clipper}				
\end{tabular}}
\caption{Five nearest neighbors of the word {\it chip} in a target and three source semantic spaces before (-) and after (+) transfer via MST+GVT configuration}
\label{tab:NN}
\end{table}

\subsection{Topics and Nearest Neighbors (NN)} 
\enote{pkj}{include detailed explanation with example from the table}
For topic level inspection, we first extract topics using the rows of ${\bf W}$ of source and target corpora. 
Table \ref{topiccoherenceexamples} shows the topics (top-5 words) from source and target domains. 
Observe that the target topics become more coherent after transfer learning (i.e., +GVT) from one or more sources. The blue color signifies that a target topic has imitated certain topic words from the source.  
We also show a topic (the last) improved due to multi-source transfer. 

For word level inspection, we extract word representations using the columns of ${\bf W}$. 
Table \ref{tab:NN} shows nearest neighbors (NNs) of the word {\it chip} in {\texttt{20NSshort}} (target) corpus, 
before (-) and after (+) topic knowledge transfer via GVT using three sources (i.e., MST+GVT).
Observe that the NNs in the target become more meaningful by gaining knowledge mainly from {\texttt{20NS}} source. 

\enote{pkj}{TODO: show additional example of topic transfer and 5-NN. May be from medical domain}
\enote{pkj}{(To Do: Include a a detailed explanation. E.g.: cosine similarity, etc)}






\enote{pkj}{Possible to do:\\
1. IR: P@10, P@0.02\\
2.IR plots for Biomedical data\\
3. Qualitative Topics for Biomedical data\\
4. Coh abalation: 50, 100 and 200 topics. top 5, 10 and 20 words. \\
5. few shot setting: addtional expriments for PPL, IR for 2 more addtional data \\
6. topic redundancy scores: JMLR paper\\
7. PPL and IR: 50 and 100 topics also.\\
8. COH: extrinsic and intrinsic\\
9. Related works section: NADE, RSM, NVDM, PRodLDA, DocNADEe, etc.\\
10. Another example qualitative analysis: +- GVT. \\
11. Another example qualitative analysis: Nearest neighbour\\
12. Unsupervised text classification?  \\
13. NVDM+GVT\\
14. Second order embeddings: Topological view (ToV) + Use wornet relation for additional edges
}


\enote{pkj}{related section including topic modeling without word embeddings and topic modeling with transfer learning i.e., using word embeddings}
 
\section{Conclusion}
We have presented a state-of-the-art neural topic modeling framework using multi-view embedding spaces: 
pretrained topic-embeddings and word-embeddings (context-sensitive and context-insensitive) from one or many sources  
to improve quality of topics and document representations.  

\section*{Acknowledgments}
This research was supported by the Federal Ministry for Economic Affairs and Energy (Bundeswirtschaftsministerium: \href{bmwi.de}{bmwi.de}), grant 01MD19003E (PLASS: Platform for Analytical Supply Chain Mangement Services, \href{plass.io}{plass.io}) at Siemens AG (Technology- Machine Intelligence), Munich Germany.

\bibliographystyle{acl_natbib}
\bibliography{emnlp2020}

\begin{thebibliography}{25}
\expandafter\ifx\csname natexlab\endcsname\relax\def\natexlab#1{#1}\fi

\bibitem[{Beltagy et~al.(2019)Beltagy, Lo, and Cohan}]{scibert2019}
Iz~Beltagy, Kyle Lo, and Arman Cohan. 2019.
\newblock \href {https://www.aclweb.org/anthology/D19-1371} {{S}ci{BERT}: A
  pretrained language model for scientific text}.
\newblock In \emph{Proceedings of the 2019 Conference on Empirical Methods in
  Natural Language Processing and the 9th International Joint Conference on
  Natural Language Processing (EMNLP-IJCNLP)}, Hong Kong, China. Association
  for Computational Linguistics.

\bibitem[{Bengio et~al.(2003)Bengio, Ducharme, Vincent, and
  Janvin}]{DBLP:journals/jmlr/BengioDVJ03}
Yoshua Bengio, R{\'{e}}jean Ducharme, Pascal Vincent, and Christian Janvin.
  2003.
\newblock \href {http://www.jmlr.org/papers/v3/bengio03a.html} {A neural
  probabilistic language model}.
\newblock \emph{Journal of Machine Learning Research}, 3:1137--1155.

\bibitem[{Blei et~al.(2003)Blei, Ng, and Jordan}]{DBLP:journals/jmlr/BleiNJ03}
David~M. Blei, Andrew~Y. Ng, and Michael~I. Jordan. 2003.
\newblock \href {http://www.jmlr.org/papers/v3/blei03a.html} {Latent dirichlet
  allocation}.
\newblock \emph{Journal of Machine Learning Research}, 3:993--1022.

\bibitem[{Bojanowski et~al.(2017)Bojanowski, Grave, Joulin, and
  Mikolov}]{DBLP:journals/tacl/BojanowskiGJM17}
Piotr Bojanowski, Edouard Grave, Armand Joulin, and Tomas Mikolov. 2017.
\newblock Enriching word vectors with subword information.
\newblock \emph{{TACL}}, 5:135--146.

\bibitem[{Cao et~al.(2010)Cao, Pan, Zhang, Yeung, and
  Yang}]{DBLP:conf/aaai/CaoPZYY10}
Bin Cao, Sinno~Jialin Pan, Yu~Zhang, Dit{-}Yan Yeung, and Qiang Yang. 2010.
\newblock \href {http://www.aaai.org/ocs/index.php/AAAI/AAAI10/paper/view/1823}
  {Adaptive transfer learning}.
\newblock In \emph{Proceedings of the Twenty-Fourth {AAAI} Conference on
  Artificial Intelligence, {AAAI} 2010, Atlanta, Georgia, USA, July 11-15,
  2010}. {AAAI} Press.

\bibitem[{Chang et~al.(2009)Chang, Boyd{-}Graber, Gerrish, Wang, and
  Blei}]{DBLP:conf/nips/ChangBGWB09}
Jonathan Chang, Jordan~L. Boyd{-}Graber, Sean Gerrish, Chong Wang, and David~M.
  Blei. 2009.
\newblock Reading tea leaves: How humans interpret topic models.
\newblock In \emph{Advances in Neural Information Processing Systems 22: 23rd
  Annual Conference on Neural Information Processing Systems 2009. Proceedings
  of a meeting held 7-10 December 2009, Vancouver, British Columbia, Canada.},
  pages 288--296.

\bibitem[{Das et~al.(2015)Das, Zaheer, and Dyer}]{P15-1077}
Rajarshi Das, Manzil Zaheer, and Chris Dyer. 2015.
\newblock \href {https://doi.org/10.3115/v1/P15-1077} {Gaussian lda for topic
  models with word embeddings}.
\newblock In \emph{Proceedings of the 53rd Annual Meeting of the Association
  for Computational Linguistics and the 7th International Joint Conference on
  Natural Language Processing (Volume 1: Long Papers)}, pages 795--804.
  Association for Computational Linguistics.

\bibitem[{Devlin et~al.(2019)Devlin, Chang, Lee, and
  Toutanova}]{BERT:DevlinCLT19}
Jacob Devlin, Ming{-}Wei Chang, Kenton Lee, and Kristina Toutanova. 2019.
\newblock \href {https://doi.org/10.18653/v1/n19-1423} {{BERT:} pre-training of
  deep bidirectional transformers for language understanding}.
\newblock In \emph{Proceedings of the 2019 Conference of the North American
  Chapter of the Association for Computational Linguistics: Human Language
  Technologies, {NAACL-HLT} 2019, Minneapolis, MN, USA, June 2-7, 2019, Volume
  1 (Long and Short Papers)}, pages 4171--4186. Association for Computational
  Linguistics.

\bibitem[{Gupta et~al.(2019{\natexlab{a}})Gupta, Chaudhary, Buettner, and
  Sch{\"u}tze}]{pankajgupta:2019iDocNADEe}
Pankaj Gupta, Yatin Chaudhary, Florian Buettner, and Hinrich Sch{\"u}tze.
  2019{\natexlab{a}}.
\newblock \href {{http://arxiv.org/abs/1809.06709}} {Document informed neural
  autoregressive topic models with distributional prior}.
\newblock In \emph{Proceedings of the Thirty-Third AAAI Conference on
  Artificial Intelligence}.

\bibitem[{Gupta et~al.(2019{\natexlab{b}})Gupta, Chaudhary, Buettner, and
  Sch{\"{u}}tze}]{GuptatexttovecICLR2019}
Pankaj Gupta, Yatin Chaudhary, Florian Buettner, and Hinrich Sch{\"{u}}tze.
  2019{\natexlab{b}}.
\newblock \href {https://openreview.net/forum?id=rkgoyn09KQ} {{textTOvec}: Deep
  contextualized neural autoregressive topic models of language with
  distributed compositional prior}.
\newblock In \emph{7th International Conference on Learning Representations,
  {ICLR} 2019, New Orleans, LA, USA, May 6-9, 2019}. OpenReview.net.

\bibitem[{Gupta et~al.(2020)Gupta, Chaudhary, Runkler, and
  Sch{\"{u}}tze}]{Guptalifelongicml2020}
Pankaj Gupta, Yatin Chaudhary, Thomas~A. Runkler, and Hinrich Sch{\"{u}}tze.
  2020.
\newblock \href {http://proceedings.mlr.press/v119/gupta20a.html} {Neural topic
  modeling with continual lifelong learning}.
\newblock In \emph{Proceedings of the 37th International Conference on Machine
  Learning, {ICML} 2020, 13-18 July 2020, Virtual Event}, volume 119 of
  \emph{Proceedings of Machine Learning Research}, pages 3907--3917. {PMLR}.

\bibitem[{Larochelle and Lauly(2012)}]{DBLP:conf/nips/LarochelleL12}
Hugo Larochelle and Stanislas Lauly. 2012.
\newblock \href
  {http://papers.nips.cc/paper/4613-a-neural-autoregressive-topic-model} {A
  neural autoregressive topic model}.
\newblock In \emph{Advances in Neural Information Processing Systems 25: 26th
  Annual Conference on Neural Information Processing Systems}, pages
  2717--2725.

\bibitem[{Larochelle and Murray(2011)}]{DBLP:journals/jmlr/LarochelleM11}
Hugo Larochelle and Iain Murray. 2011.
\newblock The neural autoregressive distribution estimator.
\newblock In \emph{Proceedings of the Fourteenth International Conference on
  Artificial Intelligence and Statistics, AISTATS}, volume~15 of \emph{{JMLR}
  Proceedings}, pages 29--37. JMLR.org.

\bibitem[{Le and Mikolov(2014)}]{DBLP:conf/icml/LeM14}
Quoc~V. Le and Tomas Mikolov. 2014.
\newblock \href {http://jmlr.org/proceedings/papers/v32/le14.html} {Distributed
  representations of sentences and documents}.
\newblock In \emph{Proceedings of the 31th International Conference on Machine
  Learning, {ICML}}, volume~32 of \emph{{JMLR} Workshop and Conference
  Proceedings}, pages 1188--1196. JMLR.org.

\bibitem[{Liu et~al.(2019)Liu, Ott, Goyal, Du, Joshi, Chen, Levy, Lewis,
  Zettlemoyer, and Stoyanov}]{RoBERTa2019}
Yinhan Liu, Myle Ott, Naman Goyal, Jingfei Du, Mandar Joshi, Danqi Chen, Omer
  Levy, Mike Lewis, Luke Zettlemoyer, and Veselin Stoyanov. 2019.
\newblock \href {http://arxiv.org/abs/1907.11692} {Roberta: {A} robustly
  optimized {BERT} pretraining approach}.
\newblock \emph{CoRR}, abs/1907.11692.

\bibitem[{Miao et~al.(2016)Miao, Yu, and Blunsom}]{DBLP:conf/icml/MiaoYB16}
Yishu Miao, Lei Yu, and Phil Blunsom. 2016.
\newblock Neural variational inference for text processing.
\newblock In \emph{Proceedings of the 33nd International Conference on Machine
  Learning, {ICML} 2016, New York City, NY, USA, June 19-24, 2016}, volume~48
  of \emph{{JMLR} Workshop and Conference Proceedings}, pages 1727--1736.
  JMLR.org.

\bibitem[{Mikolov et~al.(2013{\natexlab{a}})Mikolov, Chen, Corrado, and
  Dean}]{Mikolov:word2vec2013}
Tomas Mikolov, Kai Chen, Greg Corrado, and Jeffrey Dean. 2013{\natexlab{a}}.
\newblock \href {http://arxiv.org/abs/1301.3781} {Efficient estimation of word
  representations in vector space}.
\newblock In \emph{1st International Conference on Learning Representations,
  {ICLR} 2013, Scottsdale, Arizona, USA, May 2-4, 2013, Workshop Track
  Proceedings}.

\bibitem[{Mikolov et~al.(2013{\natexlab{b}})Mikolov, Sutskever, Chen, Corrado,
  and Dean}]{DBLP:conf/nips/MikolovSCCD13}
Tomas Mikolov, Ilya Sutskever, Kai Chen, Gregory~S. Corrado, and Jeffrey Dean.
  2013{\natexlab{b}}.
\newblock \href
  {http://papers.nips.cc/paper/5021-distributed-representations-of-words-and-phrases-and-their-compositionality}
  {Distributed representations of words and phrases and their
  compositionality}.
\newblock In \emph{Advances in Neural Information Processing Systems 26: 27th
  Annual Conference on Neural Information Processing Systems}, pages
  3111--3119.

\bibitem[{Moen and Ananiadou(2013)}]{moen2013distributional}
SPFGH Moen and Tapio Salakoski2~Sophia Ananiadou. 2013.
\newblock Distributional semantics resources for biomedical text processing.
\newblock \emph{Proceedings of LBM}, pages 39--44.

\bibitem[{Nguyen et~al.(2015)Nguyen, Billingsley, Du, and
  Johnson}]{DBLP:journals/tacl/NguyenBDJ15}
Dat~Quoc Nguyen, Richard Billingsley, Lan Du, and Mark Johnson. 2015.
\newblock \href
  {https://tacl2013.cs.columbia.edu/ojs/index.php/tacl/article/view/582}
  {Improving topic models with latent feature word representations}.
\newblock \emph{{TACL}}, 3:299--313.

\bibitem[{Pennington et~al.(2014)Pennington, Socher, and Manning}]{D14-1162}
Jeffrey Pennington, Richard Socher, and Christopher Manning. 2014.
\newblock \href {https://doi.org/10.3115/v1/D14-1162} {Glove: Global vectors
  for word representation}.
\newblock In \emph{Proceedings of the 2014 Conference on Empirical Methods in
  Natural Language Processing (EMNLP)}, pages 1532--1543. Association for
  Computational Linguistics.

\bibitem[{Peters et~al.(2018)Peters, Neumann, Iyyer, Gardner, Clark, Lee, and
  Zettlemoyer}]{N18-1202}
Matthew Peters, Mark Neumann, Mohit Iyyer, Matt Gardner, Christopher Clark,
  Kenton Lee, and Luke Zettlemoyer. 2018.
\newblock \href {https://doi.org/10.18653/v1/N18-1202} {Deep contextualized
  word representations}.
\newblock In \emph{Proceedings of the 2018 Conference of the North American
  Chapter of the Association for Computational Linguistics: Human Language
  Technologies, Volume 1 (Long Papers)}, pages 2227--2237. Association for
  Computational Linguistics.

\bibitem[{R{\"{o}}der et~al.(2015)R{\"{o}}der, Both, and
  Hinneburg}]{DBLP:conf/wsdm/RoderBH15}
Michael R{\"{o}}der, Andreas Both, and Alexander Hinneburg. 2015.
\newblock \href {https://doi.org/10.1145/2684822.2685324} {Exploring the space
  of topic coherence measures}.
\newblock In \emph{Proceedings of the Eighth {ACM} International Conference on
  Web Search and Data Mining, {WSDM} 2015, Shanghai, China, February 2-6,
  2015}, pages 399--408. {ACM}.

\bibitem[{Salakhutdinov and Hinton(2009)}]{DBLP:conf/nips/SalakhutdinovH09}
Ruslan Salakhutdinov and Geoffrey~E. Hinton. 2009.
\newblock \href
  {http://papers.nips.cc/paper/3856-replicated-softmax-an-undirected-topic-model}
  {Replicated softmax: an undirected topic model}.
\newblock In \emph{Advances in Neural Information Processing Systems 22: 23rd
  Annual Conference on Neural Information Processing Systems}, pages
  1607--1614. Curran Associates, Inc.

\bibitem[{Srivastava and Sutton(2017)}]{SrivastavaSutton}
Akash Srivastava and Charles Sutton. 2017.
\newblock \href {https://arxiv.org/pdf/1703.01488.pdf} {Autoencoding
  variational inference for topic models}.
\newblock In \emph{5th International Conference on Learning Representations,
  ICLR}.

\end{thebibliography}

\appendix

\section{Data Description}\label{secappendix:datadescription}
In order to evaluate knowledge transfer within unsupervised neural topic modeling, 
we use the following seven datasets in the target domain $\mathcal{T}$ following the similar experimental setup as in DocNADEe: 
(1) \texttt{20NSshort}: We take documents from 20NewsGroups data, with document size (number of words) less than 20.
(2)  \texttt{20NSsmall}: We sample 20 document (each having more than $200$ words) for training from each class of the 20NS dataset. For validation and test, 10 document for each class. Therefore, it is a corpus of few (long) documents. 
(3) \texttt{TMNtitle}: Titles of the Tag My News (TMN) news dataset. 
(4) \texttt{R21578title}:  Reuters corpus, a collection of new stories from \url{nltk.corpus}. We take titles of the documents.
(5) \texttt{Ohsumedtitle}: Titles of Ohsumed abstracts. Source: \url{disi.unitn.it/moschitti/corpora.htm}.
(6) \texttt{Ohsumed}: Ohsumed dataset, collection of medical abstracts. Source: \url{disi.unitn.it/moschitti/corpora.htm}.
(7) \texttt{TMN}: The Tag My News (TMN) news dataset. 

To {\bf prepare knowledge base} of word embedings (local semantics) and latent topics (global semantics) features, 
we use the following six datasets in the source ${\mathcal S}$: 
(1)  \texttt{20NS}: 20NewsGroups corpus, a collection of  news stories from \url{nltk.corpus}.  
(2) \texttt{TMN}: The Tag My News (TMN) news dataset. 
(3) \texttt{R21578}: Reuters corpus, a collection of new stories from \url{nltk.corpus}.   
(4) \texttt{AGnews}:  AGnews data sellection. 
 \texttt{PubMed}: Medical abstracts of randomized controlled trials. Source: \url{https://github.com/Franck-Dernoncourt/pubmed-rct}.  

See Table 3 (in paper content) describes each of the datasets, where a short-text refers to a text document having less than 15 words.
Notice that each of the datasets in the target and source domains, we see overlap in their label spaces. 
See Table 4 for the label information for each of the source and target corpora. 
Additionally in supplementary, we have also provided the code and pre-processed datasets used in our experiments. 

\section{Getting Word and Latent Topic Representations from Source(s)}\label{secappendix:latenttopicgeneration}

Since in DocNADE, the column of ${\bf W}_{:,v_i}$ gives a word vector of the word $v_i$, 
therefore the dimension of word embeddings in each of the ${\bf E}^k$ is same (i.e., $H=200$). 
Thus, we prepare the knowledge base of word representations ${\bf E}^k$ from $k$th source using DocNADE, 
where each word vector is of $H=200$ dimension.  
 
Since the row vector of ${\bf W}_{j,:}$ in DocNADE  encodes $j$th topic feature, therefore 
each latent topic (i.e., row) in feature matrix ${\bf W}$ is a vector of $K$ dimension, 
corresponding the definition of topics that it is a distribution over vocabulary. 
$H$ is the number of latent topics and $K$ is the vocabulary size, where $K$ varies across corpora. 
Thus, we train DocNADE to learn a feature matrix specific to each of the source corpora, 
e.g.  ${\bf W}^{k} \in \mathbb{R}^{H \times K}$ of $k$th source. 

For a target corpus of vocabulary size $K^{'}$, the DocNADE learns a feature matrix ${\bf W}^{\mathcal T} \in \mathbb{R}^{H \times K'}$.  
Similarly, ${\bf W}^{k} \in \mathbb{R}^{H \times K}$ for $k$th source of vocabulary size $K$. 
Since in the sparse-data setting for the target,  $K' << K$ due to additional word in the source. 
To perform GVT, we need the same topic feature dimensions in the target and source, i.e., $K'$ of the target. 
Therefore, we remove those column vectors from ${\bf W}^{k} \in \mathbb{R}^{H \times K}$ of the $k$th source for which  
there is no corresponding word in the vocabulary of the target domain.  
As a result, we obtain ${\bf Z}^{k}$ as a latent topic feature matrix to be used in knowledge transfer to the target domain.  
Following the similar steps, we prepare a KB of ${\bf Z}$s such that each latent topic feature matrix 
from a source domain gets the same topic feature dimension as the target.

\begin{table}[t]
      \centering
        \resizebox{.48\textwidth}{!}{
       \begin{tabular}{c|c}
        {\bf data}    &    {\bf labels / classes } \\ \hline 
	\multirow{1}{*}{TMN*}    &   world, us, sport, business, sci$\_$tech, entertainment, health \\  \hline
	\multirow{1}{*}{AGnews}    &   business, sci$\_$tech, sports, world \\  \hline

										   &   misc.forsale, comp.graphics, rec.autos, comp.windows.x, \\ 
   				20NS					 &   rec.sport.baseball, sci.space, rec.sport.hockey, \\ 
                     20NSshort,                                 				& soc.religion.christian, rec.motorcycles, comp.sys.mac.hardware,\\
			20NSsmall,  							& talk.religion.misc, sci.electronics, comp.os.ms-windows.misc,\\ 
                                                     				 & sci.med, comp.sys.ibm.pc.hardware, talk.politics.mideast,\\
										&  talk.politics.guns, talk.politics.misc, alt.atheism, sci.crypt\\ \hline		

       								   &   trade, grain, crude, corn, rice, rubber, sugar, palm-oil, \\
									& veg-oil, ship, coffee, wheat, gold, acq, interest, money-fx,\\
									& carcass, livestock, oilseed, soybean, earn, bop, gas, lead, zinc,\\
				R21578title 					&  gnp, soy-oil, dlr, yen, nickel, groundnut, heat, sorghum, sunseed,  \\
				R21578					& cocoa, rapeseed, cotton, money-supply, iron-steel, palladium, \\
									& platinum, strategic-metal, reserves, groundnut-oil, lin-oil, meal-feed, \\
									& sun-meal, sun-oil, hog, barley, potato, orange, soy-meal, cotton-oil,  \\
									& fuel, silver, income, wpi, tea, lei, coconut, coconut-oil, copra-cake,  \\
									& propane, instal-debt, nzdlr, housing, nkr, rye, castor-oil, palmkernel, \\ 
									& tin, copper, cpi,   pet-chem, 	 rape-oil, oat, naphtha, cpu, rand, alum 		\\ \hline
							
         \end{tabular}}
         \caption{Label space of the corpora. TMN*:TMN or TMNtitle}
\label{tab:labelspace}
\end{table}

\begin{table}[t]
      \centering
      \resizebox{.40\textwidth}{!}{
        \begin{tabular}{c|c}
         \hline 
        {\bf Hyperparameter}               & {\bf Search Space} \\ \hline
          retrieval fraction        &    [0.02]                        \\
          learning rate        &    [{0.001}]                       \\
          hidden units, $H$         &      [200]               \\ 
          activation function ($g$)        &      {sigmoid}     \\
          iterations        &      [100]      \\
          $\lambda^k$        &      [1.0, 0.5, 0.1]  \\ 
          $\gamma^k$        &      [0.1, 0.01, 0.001]  \\ \hline 
         \end{tabular}}
          \caption{Hyperparameters in Generalization experiments of DocNADE, DocNADEe, LVT, GVT and MVT}\label{HyperparametersinGeneralization}
 \end{table}%

\begin{table}[t]
      \centering
        \resizebox{.40\textwidth}{!}{
       \begin{tabular}{c|c}
        \hline 
        {\bf Hyperparameter}               & {\bf Search Space} \\ \hline
          retrieval fraction        &    [0.02]                        \\
          learning rate        &    [{0.001}]                       \\
          hidden units, $H$         &      [200]               \\ 
          activation function ($g$)        &      {tanh}     \\
          iterations        &      [100]      \\
          $\lambda^k$        &      [1.0, 0.5, 0.1]  \\ 
          $\gamma^k$        &      [0.1, 0.01, 0.001]  \\ \hline 
         \end{tabular}}
         \caption{Hyperparameters search in the IR task, 
where $\lambda^k$ and $\gamma^k$ are weights for $k$th source.}
\label{appendixHyperparametersinIR}
\end{table}

\begin{table}[t]
\center
\renewcommand*{\arraystretch}{1.}
\resizebox{.48\textwidth}{!}{
\setlength\tabcolsep{3.5pt}
\begin{tabular}{r|r|r|ccc|ccc|ccc}
\multirow{3}{*}{\bf }  & \multirow{1}{*}{\bf }   &     \multirow{1}{*}{\bf } & \multicolumn{9}{c}{{\bf Scores on Target Corpus} ({\it in sparse-data  setting})}   \\ \cline{3-12}
     &   \multicolumn{2}{c|}{\bf } &   \multicolumn{3}{c|}{\texttt{20NSshort}}     &  \multicolumn{3}{c|}{\texttt{TMNtitle}}   &    \multicolumn{3}{c}{\texttt{20NSsmall}} \\
    &   \multicolumn{2}{c|}{\bf Type}  &   $PPL$  &  $COH$ &  $IR$ &    $PPL$  &  $COH$ &  $IR$  & $PPL$  &  $COH$ &  $IR$  \\ \hline 

\multirow{4}{*}{$+$} & \multirow{4}{*}{\it MST}  &  LVT      		&  667   & .661   & .308      &  670   & .730   & .535     	     &  610   & .440   & .286  			 \\
 &					   &  GVT         &  651   & .658   & .285        &  701   & .712   & .523     	   &  602   & .460   & {.273} 			\\ 
 &					  &  MVT    	&  667   & .660   & {.309}     & 667    & {.730}   & {.535} 	&  608   & {.441}   & .293   			   \\
 &					  &  + Glove     &  662   & .677   & .296         &  672   & {.731}   & .540         &  634   & .412   & .207   		\\ \hline \hline 

\multirow{4}{*}{$\times$} & \multirow{4}{*}{\it MST}  	   &  LVT      	&  {\bf 640}   & .678   & .308     &  {\bf 663}   & .732   & .547     		     &  {\bf 596}   & .442   & .277   			 \\
 &					   &  GVT          &  658   & .705   & .305     &  704   & .746   & .550     		     &  599   & .585   & {\bf .326} 		  \\ 
 &					  &  MVT    	&  656   & {\bf .721}   	& {\bf .314}     &  680   & {\bf .752}   & {\bf .556}     	& 600	  & {\bf .600}   & {\bf .285}      \\
 &					  &  + Glove     &  644   & .719   & .293     &  687   & {\bf .752}   & .538            &  609   & .586   & .282   			
\end{tabular}}
\caption{\{$\lambda$, $\gamma$\} as Parameter ($+$) vs Hyperparameters ($\times$): Perplexity (PPL), topic coherence (COH) and precision@recall (IR) at retrieval fraction $0.02$, when $\lambda$ and $\gamma$ are (1) learned with backpropagation,
 and (2) treated as hyperparameters.  Results suggest the superiority of the second configuration.  
}
\label{tab:scoressupplementary}
\end{table}

\section{Experimental Setup}\label{secappendix:exprimentalsetup} 

For DocNADE and DocNADEe in different knowledge transfer configurations, 
we follow the same experimental setup as in DocNADE and DocNADEe. 
We rerun DocNADE and DocNADEe using the code released for DocNADEe. 
For all the hyperparameters, optimal values are selected based on the performance on development set.

\subsection{Experimental Setup for Generalization}\label{secappendix:exprimentalsetupPPL}

We set the maximum number of training passes to 100, topics to 200 and the learning rate to 0.001 with {\it sigmoid} hidden activation.  
Since the baseline DocNADE and DocNADEe reported better scores in PPL for $H=200$ topics than using $50$, therefore we use $H=200$ in our experiments. 
See Table \ref{HyperparametersinGeneralization}  for hyperparameters used in generalization task, i.e., computing PPL. 

\subsection{Experimental Setup for IR Task}\label{secappendix:exprimentalsetupIR}
We treat all test documents as queries to retrieve a fraction of the closest documents in the original training set using cosine similarity  between their document vectors. To compute retrieval precision for each fraction (e.g., $0.02$),  
we average the number of retrieved training documents with the same label as the query. 

We set the maximum number of training passes to 100, topics to 200 and the learning rate to 0.001 with {\it tanh} hidden activation.  
Since the baseline DocNADE and DocNADEe reported better scores in precision for the retrieval task for $H=200$ topics than using $50$, 
therefore we use $H=200$ in our experiments. 
We follow the similar experimental setup as in DocNADEe.
For model selection, we used the validation set as the query set and used the average precision at 0.02 retrieved documents as the
performance measure.  
Note that the labels are not used during training. The class labels are only used to check if the retrieved documents have the same class label as the query document. 
To perform document retrieval, we use the same train/development/test split of documents as for PPL setup.   

Given DocNADE, the representation of a document of size $D$ can be computed by taking the last hidden vector ${\bf h}_D$ 
at the autoregressive step $D$.  Since, the RSM and DocNADE strictly outperformed LDA, therefore we only compare DocNADE and its recent 
 extension DocNADEe. We use the same number of topic dimensions ($H=200$) across all the source and target in training DocNADE. 
 
See Table \ref{appendixHyperparametersinIR} for the hyperparameters in the document retrieval task, 
where $\lambda^k$ and $\gamma^k$ are weights for $k$th source. 
We use the same grid-search for all the source domains. We set  $\gamma^k$ smaller than $\lambda^k$ to control 
the degree of imitation of the source domain(s) by the target domain.
We use the development set of the target corpus to find the 
optimal setting in different configurations of knowledge transfers from several sources. 
\subsection{\{$\lambda$, $\gamma$\} as Parameter vs Hyperparameters}\label{secappendix:paramvshyperams}

Here, we treat $\lambda$ and $\gamma$ as parameters of the model, instead of hyperparameters and learn them with backpropagation. 
We initialize each $\lambda^k$  = $0.5$ and $\gamma^k$  = $0.01$ for each of the sources. 
We perform experiments on short-text datasets in MST+LVT, MST+GVT and MST+MVT configurations. 
We evaluate the topic modeling using PPL, topic coherence and retrieval accuracy.  
Table \ref{tab:scoressupplementary} reports the scores,  when $\lambda$ and $\gamma$ are (1) learned with backpropagation,
 and (2) treated as hyperparameters.  The experimental results suggest that the second configuration performs better the former. 
Thus,  we have reported scores considering  \{$\lambda$, $\gamma$\} as hyperparameters.

\subsection{Reproducibility: Optimal Configurations of $\lambda$ and $\gamma$}\label{secappendix:Reproducibility}

As mentioned in Tables \ref{HyperparametersinGeneralization} and \ref{appendixHyperparametersinIR}, 
the hyper-parameter $\lambda^k$ takes on values in [1.0, 0.5, 0.1]  for each of the word embeddings matrix ${\bf E}^k$ and 
$\gamma^k$ in [0.1, 0.01, 0.001] for each of the latent topic features ${\bf Z}^k$, respectively for the $k^{th}$ source domain.
To determine an optimal configuration, we perform grid-search over the values and use the scores on the 
development set to determine the best setting.  We have a common model for PPL and COH scores due to generalization. 

To {\bf reproduce} scores (best/bold in Table 5,  
we mentioned the best settings of ($\lambda^k$, $\gamma^k$) in MST+MVT configuration for each of the target and source combinations:
\begin{enumerate}
\item Generalization (PPL and COH) in MST+MVT when {\bf target} is \texttt{20NSshort}:  
($\lambda^{20NS}= 1.0$, $\gamma^{20NS} = 0.001$, $\lambda^{TMN} = 0.1$, $\gamma^{TMN}= 0.001$, $\lambda^{R21578} = 0.5$, $\gamma^{R21578} = 0.001$, $\lambda^{AGnews} = 0.1$, $\gamma^{AGnews} = 0.001$

\item Generalization (PPL and COH) in MST+MVT when {\bf target} is \texttt{TMNtitle}:  
($\lambda^{20NS}= 0.1$, $\gamma^{20NS} = 0.001$, $\lambda^{TMN} = 1.0$, $\gamma^{TMN}= 0.001$, $\lambda^{R21578} = 0.5$, $\gamma^{R21578} = 0.001$, $\lambda^{AGnews} = 1.0$, $\gamma^{AGnews} =0.001$

\item Generalization (PPL and COH) in MST+MVT when {\bf target} is \texttt{R21578title}:  
($\lambda^{20NS}= 0.1$, $\gamma^{20NS} = 0.001$, $\lambda^{TMN} = 0.5$, $\gamma^{TMN}= 0.001$, $\lambda^{R21578} = 1.0$, $\gamma^{R21578} = 0.001$, $\lambda^{AGnews} = 1.0$, $\gamma^{AGnews} = 0.001$

\item Generalization (PPL and COH) in MST+MVT when {\bf target} is \texttt{20NSsmall}:  
($\lambda^{20NS}= 0.5$, $\gamma^{20NS} = 0.001$, $\lambda^{TMN} = 0.1$, $\gamma^{TMN}= 0.001$, $\lambda^{R21578} = 0.1$, $\gamma^{R21578} = 0.001$, $\lambda^{AGnews} = 0.1$, $\gamma^{AGnews} = 0.001$

\item Generalization (PPL and COH) in MST+MVT when {\bf target} is \texttt{Ohsumedtitle}:  
($\lambda^{AGnews}= 0.1$, $\gamma^{AGnews} = 0.001$, $\lambda^{PubMed} = 1.0$, $\gamma^{PubMed}= 0.001$

\item Generalization (PPL and COH) in MST+MVT when {\bf target} is \texttt{Ohsumed}:  
($\lambda^{AGnews}= 0.1$, $\gamma^{AGnews} = 0.001$, $\lambda^{PubMed} = 1.0$, $\gamma^{PubMed}= 0.001$

\item IR  in MST+MVT when {\bf target} is \texttt{20NSshort}:  
($\lambda^{20NS}= 1.0$, $\gamma^{20NS} = 0.1$, $\lambda^{TMN} = 0.5$, $\gamma^{TMN}= 0.01$, $\lambda^{R21578} = 0.1$, $\gamma^{R21578} = 0.001$, $\lambda^{AGnews} = 1.0$, $\gamma^{AGnews} = 0.01$

\item IR in MST+MVT when {\bf target} is \texttt{TMNtitle}:  
($\lambda^{20NS}= 0.1$, $\gamma^{20NS} = 0.01$, $\lambda^{TMN} = 1.0$, $\gamma^{TMN}= 0.01$, $\lambda^{R21578} = 0.1$, $\gamma^{R21578} = 0.01$, $\lambda^{AGnews} = 0.5$, $\gamma^{AGnews} = 0.001$

\item IR in MST+MVT when {\bf target} is \texttt{R21578title}:  
($\lambda^{20NS}= 0.1$, $\gamma^{20NS} = 0.01$, $\lambda^{TMN} = 1.0$, $\gamma^{TMN}= 0.01$, $\lambda^{R21578} = 1.0$, $\gamma^{R21578} = 0.01$, $\lambda^{AGnews} = 0.5$, $\gamma^{AGnews} = 0.001$

\item IR in MST+GVT when {\bf target} is \texttt{20NSsmall}:  
($\gamma^{20NS} = 0.01$,  $\gamma^{TMN}= 0.01$, $\gamma^{R21578} = 0.1$, $\gamma^{AGnews} = 0.01$

\item IR in MST+MVT when {\bf target} is \texttt{Ohsumedtitle}:  
($\lambda^{AGnews}= 0.1$, $\gamma^{AGnews} = 0.001$, $\lambda^{PubMed} = 1.0$, $\gamma^{PubMed}= 0.1$

\item IR in MST+MVT when {\bf target} is \texttt{Ohsumed}:  
($\lambda^{AGnews}= 0.1$, $\gamma^{AGnews} = 0.001$, $\lambda^{PubMed} = 0.5$, $\gamma^{PubMed}= 0.1$

\end{enumerate}

The hyper-parameters mentioned above also applies to a single source transfer configuration.

\end{document}